\documentclass[11pt,english]{article}
\usepackage[sc]{mathpazo}
\usepackage{geometry}
\usepackage{color}
\usepackage{array}
\usepackage{bbding}
\usepackage{multirow}
\usepackage[fleqn]{amsmath}
\usepackage{amssymb}
\usepackage{amsthm}
\usepackage{graphicx}
\usepackage{natbib}
\usepackage{lscape}
\usepackage[hidelinks]{hyperref}
\usepackage{comment}
\usepackage{bm}
\usepackage[title]{appendix}
\usepackage{longtable}
\usepackage{mathtools}
\usepackage{chngcntr}
\usepackage{authblk}
\usepackage{siunitx}
\usepackage{babel}
\usepackage{dsfont}
\usepackage[most]{tcolorbox}
\usepackage[short]{optidef}

\makeatletter
\allowdisplaybreaks
\usepackage[mathlines, pagewise]{lineno}
\usepackage{etoolbox} 

\newcommand*\linenomathpatchAMS[1]{%
	\expandafter\pretocmd\csname #1\endcsname {\linenomathAMS}{}{}%
	\expandafter\pretocmd\csname #1*\endcsname{\linenomathAMS}{}{}%
	\expandafter\apptocmd\csname end#1\endcsname {\endlinenomath}{}{}%
	\expandafter\apptocmd\csname end#1*\endcsname{\endlinenomath}{}{}%
}

\expandafter\ifx\linenomath\linenomathWithnumbers
\let\linenomathAMS\linenomathWithnumbers
\patchcmd\linenomathAMS{\advance\postdisplaypenalty\linenopenalty}{}{}{}
\else
\let\linenomathAMS\linenomathNonumbers
\fi

\linenomathpatchAMS{gather}
\linenomathpatchAMS{multline}
\linenomathpatchAMS{align}
\linenomathpatchAMS{alignat}
\linenomathpatchAMS{flalign}

\@ifundefined{showcaptionsetup}{}{%
 \PassOptionsToPackage{caption=false}{subfig}}
\usepackage{subfig}
\makeatother

\usepackage{babel}

\usepackage{cleveref}
\begin{document}
	
	\title{Toward LLM-Agent-Based Modeling of Transportation Systems: A Conceptual Framework}
	\author[1]{Tianming Liu}
    \author[2]{Jirong Yang}
	\author[1]{Yafeng Yin\footnote{Corresponding author. E-mail address: \textcolor{blue}{yafeng@umich.edu} (Y. Yin).}}

	\affil[1]{\small\emph{Department of Civil and Environmental Engineering, University of Michigan, Ann Arbor, United States}\normalsize}
 \affil[2]{\small\emph{Department of Electrical Engineering and Computer Science, University of Michigan, Ann Arbor, United States}\normalsize}
	\date{\today}
	\maketitle

\begin{abstract}
     In transportation system planning, agent-based models (ABMs) and microsimulations have become pivotal tools for understanding system dynamics and supporting decision-making. However, existing ABMs remain limited in their behavioral representation, flexibility, and reliance on comprehensive input data. This paper argues that recent advancements in large language models (LLMs) present a promising new avenue for enhancing agent-based transportation modeling. We propose an LLM-agent-based framework in which LLM agents act as behaviorally rich proxies for human travelers. By leveraging LLMs’ capabilities in natural language understanding, contextual reasoning, and generalization, our framework aims to overcome key limitations of traditional ABMs and unlock new modeling possibilities. We design LLM agents with structured profiles, memory systems, perception, decision-making, and action modules to align with the principles of activity-based travel demand modeling. Through system design and literature synthesis, we outline the conceptual and practical advantages of this approach and support our vision with a small-scale proof-of-concept simulation. Lastly, we discuss the remaining challenges and propose hybrid modeling as a near-term integration strategy. By positioning LLM agents as a novel and promising paradigm, we aim to expand the role of LLMs in agent-based transportation modeling and pave the way for new approaches to travel demand modeling.
\end{abstract}

\hfill\break%
\noindent\textit{Keywords} -Transportation system modeling, travel behavior, travel demand modeling, agent-based simulation, large language model, transportation planning \normalsize
	

\newpage

\section{Introduction}
In transportation planning, accurately modeling transportation systems and evaluating their performance is an essential task. Given the complexity of modern large-scale transportation networks, effective models must capture diverse traveler behaviors and dynamic interactions with infrastructure to predict travel patterns and assess outcomes of various planning scenarios. Agent-based models (ABMs) represent one of the state-of-the-art approaches in contemporary transportation planning. By simulating individual travelers as autonomous agents capable of interacting with each other and the environment, and making independent, informed decisions, ABMs offer a sophisticated representation of travel behavior. Unlike traditional aggregated models, such as those that follow the trip-based four-step modeling approach, ABMs overcome the limitations associated with data aggregation, enhance behavioral realism, and provide higher-resolution analyses. Additionally, they enable a broader range of assessment criteria and generate detailed, granular insights that support more effective transportation planning. In planning practice, ABMs are also gradually replacing classical four-step models in the planning workflows of some planning agencies \citep{bucci2018fhwa,wingo2023snapshot}.

However, despite significant advancement over 40 years, established agent-based modeling methods still face limitations in behavioral realism and practical applicability. First, decision-making mechanisms within these models typically rely on mathematical frameworks that necessitate \textit{a priori} behavioral assumptions, which may not fully reflect the nuanced and boundedly rational decision-making processes of travelers \citep{mahmassani1987boundedly,lou2010robust, ben2015response,van2022choice}. Second, agent-based models often require extensive local data for calibration, posing a significant obstacle to their widespread adoption in real-world applications \citep{manzo2014potentialities}. Furthermore, evaluating alternative planning scenarios, especially the effects of complex, targeted policies or emerging technologies, frequently demands substantial modifications to model structure and software specifications \citep{bastarianto2023agent}. This imposes stringent requirements on the design of the modeling framework, the resources needed to establish and maintain the framework, and the expertise of personnel conducting the evaluations, creating additional challenges for their implementation in transportation planning.

Recent advances in large language models (LLMs) offer promising new avenues to address longstanding challenges in agent-based transportation modeling. LLMs are deep generative neural networks trained on massive textual datasets to learn patterns in language, reasoning, and knowledge representation. They exhibit strong capabilities in natural language understanding, contextual reasoning, and adaptive decision-making, enabling them to interpret diverse inputs and simulate human-like behavior. When deployed as autonomous agents, LLMs can model complex, context-sensitive decision processes and dynamically adjust their behavior in response to evolving environmental and system-level conditions. Trained on a vast corpus of human-generated text, LLMs internalize patterns of thought, language, and social behavior, allowing them to potentially emulate key aspects of human cognition and decision-making. This training also supports their ability to generalize across varied behavioral contexts with minimal task-specific calibration. As a result, LLM agents are particularly well-suited to serve as digital proxies for human travelers within ABM frameworks, making them particularly valuable for transportation planning tasks that involve unstructured decisions, limited data, or complex policy contexts.

In this paper, we propose an LLM-agent-based modeling framework for transportation systems. To develop a structured and behaviorally realistic framework, we design the LLM agents and the system workflow that reflect characteristics of human travel behavior and align with the principles of activity-based travel demand modeling. We argue that this framework has the potential to enhance the expressiveness, generalization, and applicability of ABMs for transportation system modeling and simulation, and we present the existing literature to support its potential advantages. Additionally, we provide a demonstrative proof-of-concept of this system, showcasing its feasibility and desirable properties through observations. Finally, we comprehensively discuss the limitations and challenges associated with this approach, outlining key areas for improvement, and proposing future research directions to address both short-term implementation challenges and long-term research needs in LLM-agent-based transportation modeling.

The primary contribution of this paper is to articulate and demonstrate the potential of LLM agents to address critical limitations in current ABMs of transportation systems. By proposing a novel framework that integrates LLM agents with existing activity-based travel demand models and agent-based simulations for transportation systems, we present a forward-looking vision for more flexible, data-efficient, and realistic transportation simulations that could significantly facilitate future transportation planning. We position our proposed framework as a novel direction that could stimulate further research and collaboration among both researchers and practitioners in the fields of transportation engineering and artificial intelligence, ultimately aiming to foster the adoption of intelligent, accessible, and responsive LLM agents capable of realistically simulating complex travel behaviors and supporting decision-making in transportation planning.

The remainder of this paper is structured as follows: \Cref{sec:LR} provides a review of relevant literature on traditional agent-based modeling in transportation, LLMs, LLM agents, and LLM-agent-based social simulations. \Cref{sec:design} details the design of the proposed framework, including system structure, agent architecture, simulation workflow, and underlying rationale. \Cref{sec:prop} presents evidence from existing research to support the feasibility and advantages of the proposed framework. \Cref{sec:demonstration} presents a proof-of-concept of the proposed framework. \Cref{sec:challenges} discusses outstanding research challenges and proposes hybrid modeling as a near-term strategy for addressing current limitations. Finally, \Cref{sec:conclusion} summarizes the key findings, positions, and conclusions of the study.

\section{Literature Review} \label{sec:LR}
\subsection{ABMs of transportation systems} \label{subsec:LR_existing_ABM}

ABMs of transportation systems were first developed in the 1990s to integrate microsimulation of travel behaviors \citep{arentze2000albatross} and traffic flow \citep{krajzewicz2002sumo,barcelo2005dynamic}, as well as to address evolving regulations and requirements in transportation planning. Over time, ABMs have become a fundamental paradigm in the field, leading to the development of models such as TRANSSIMS \citep{smith1995transims}, MATSim \citep{w2016multi}, SimMobility \citep{adnan2016simmobility}, and Polaris \citep{auld2016polaris}. While the existing approaches span across and differentiate on multiple levels of travel behaviors and multiple resolutions of network dynamics, all ABM models share three key components: the agents, which are the travelers or households; the strategies of agents, which guide and regulate agents' behavior; and the simulated physical environment, which captures variables such as the transportation network, land use, and mobility services \citep{kagho2020agent}. These components allow ABMs to simulate travel behaviors and transportation system dynamics in detail and better support decision-making in planning processes. In line with these capabilities, planning organizations have begun integrating ABMs into practice. For instance, in the United States, as of 2023, organizations such as the North Central Texas Council of Governments \citep{wingo2023snapshot} and the Southern California Association of Governments \citep{he2022connected} have begun adopting or experimenting with agent-based modeling for regional transportation planning and policy evaluation.

However, despite significant advancements, existing transportation system ABMs still face challenges in behavioral representation and practical implementation. One key limitation is their ability to represent human travel behavior. While the specific methodologies vary across models and applications, behavioral modeling approaches in ABMs generally fall into two categories: econometric-based models and rule-based models, both of which require \textit{a priori} behavioral assumptions. Econometric models, such as discrete choice models, often rely on assumptions of rationality, stable preferences, and specific distributions of random errors to predict activity patterns, trip timing, and destination selection. However, such assumptions often fail to fully capture the bounded rationality and nuanced perceptions that influence human travelers' behavior. Rule-based models, including decision trees and heuristic-based decision frameworks, simulate travel behavior using predefined rules. While computationally efficient, these models struggle to represent complex decision-making processes and often lack flexibility, as their fixed rule sets limit adaptability across diverse travel scenarios.

Furthermore, existing frameworks require substantial data for proper calibration. Agent-based transportation models must capture multiple layers of travel behavior, including, but not limited to, car ownership, residential location choices, activity scheduling, travel mode choice, and route choice decisions. Each of these components requires extensive data collection and calibration from scratch. For instance, calibrating SimMobility requires integrating various data sources, such as demographic surveys, household travel surveys, taxi GPS records, and transit smartcard data, specific to the region being modeled \citep{adnan2016simmobility}. The considerable effort involved in data collection and the time-intensive calibration process poses significant challenges to the practical adoption of ABMs. Many metropolitan planning agencies lack the necessary data sources, financial resources, and workforce to support these demanding requirements. As a result, access to advanced ABMs remains limited for transportation planners and practitioners. According to a report in 2015 \citep{systematics2015status}, only 16\% of metropolitan planning agencies in the United States have plans to transition from conventional four-step models to more advanced approaches in the near future. Similarly, \cite{boyce2015forecasting} note that, despite their potential, ABMs have had a relatively modest impact on transportation planning practice thus far.

\subsection{Large language model}

In recent years, advances in artificial intelligence have led to the emergence of a new class of AI models: large language models (LLMs). Built upon sequence-to-sequence deep learning techniques, particularly transformers \citep{vaswani2017attention}, LLMs are designed to comprehend and generate content in human language. They are sequence-to-sequence models that can predict the probability distribution $\mathbb{P}$ of next word $w_i^{(u)}$ in a sequence given the previous words $w_{i-1}^{(u)},w_{i-2}^{(u)},...,w_{1}^{(u)}$, and sample from it to generate outputs:
\begin{equation*}
    w_i^{(u)}\sim\mathbb{P}(w_i^{(u)}|w_{i-1}^{(u)},w_{i-2}^{(u)},...,w_{1}^{(u)})
\end{equation*}

The primary advancement of LLMs over conventional purpose-built machine learning models lies in their scale, both in terms of model size and training data. LLMs often contain billions of parameters and are trained on vast and diverse text datasets, including books, articles, websites, online forums, social media, and research papers \citep{achiam2023gpt}. This extensive training enables them to generate high-quality text responses across a broad range of topics, making them more adaptable than traditional AI models, which are often limited to specialized tasks. While early LLMs, such as text-davinci \citep{brown2020language}, were primarily designed for text completion based on given input, state-of-the-art models like ChatGPT \citep{achiam2023gpt}, Gemini \citep{team2023gemini}, DeepSeek \citep{liu2024deepseek}, and Llama \citep{touvron2023llama} have evolved into sophisticated conversational agents capable of engaging in dynamic interactions and executing complex instructions.

Beyond their ability to replicate human language patterns, LLMs exhibit several critical capabilities that significantly enhance their versatility and adaptability. One of their key strengths is logical reasoning, which is facilitated by techniques like Chain-of-thought (CoT) prompting \citep{wei2022chain}. By breaking down complex problems into sequential steps, LLMs can mimic structured human reasoning, allowing for more reliable problem-solving and improved analytical depth. Additionally, they can leverage Retrieval-Augmented Generation (RAG) \citep{lewis2020retrieval} to dynamically access and synthesize information from designated sources, improving factual accuracy in generated responses. This capability enables more controlled and verifiable outputs, making LLMs particularly useful for applications requiring up-to-date or domain-specific knowledge. Furthermore, as their training datasets expand, LLMs demonstrate remarkable generalization abilities through few-shot learning \citep{brown2020language} and zero-shot learning \citep{kojima2022large}. These techniques allow them to perform unfamiliar tasks with minimal or no prior examples, adapting to new contexts with limited human guidance. The combination of these emergent properties enables LLMs to reason through complex scenarios, retrieve and validate external information, and efficiently tackle a wide range of problems.

\subsection{LLM agents}

The versatile capabilities of modern LLMs give rise to a significant paradigm in AI engineering: LLM agents. Broadly defined, an LLM agent is an autonomous system that leverages an LLM to perform tasks. The defining characteristic of these agents is autonomy, as they are typically equipped with modules for input processing, action determination, execution, and feedback, integrating LLMs with external tools to enable independent operation. The specific tasks and roles of LLM agents vary widely depending on their application. For instance, information-processing agents utilize the multi-modal understanding and natural language generation capabilities of LLMs to analyze and generate reports, such as monitoring transportation system performance \citep{ruan2024twitter,wang2024traffic}. Chatbot agents harness LLMs' natural language processing and information retrieval abilities to provide domain-specific support, including applications in public transit \citep{devunuri2024transitgpt} and transportation engineering \citep{wang2024transgpt}. Additionally, LLM agents can be designed for problem-solving tasks, such as automatic control \citep{guo2024controlagent}, by leveraging their reasoning abilities to optimize decision-making.

In LLM agent research, a growing body of work explores the potential of LLM agents as effective proxies for human behavior, simulating decision-making patterns across a range of domains. This approach builds on the idea that LLMs, trained on a vast corpus of human-generated content, can potentially internalize patterns of thought, expression, and decision-making, enabling them to approximate diverse human behaviors and support complex, human-like responses. Furthermore, their extensive parameter space and probabilistic architecture enable LLM agents to produce coherent, context-sensitive outputs that support credible role-playing in diverse contexts and backgrounds. As a result, LLM agents have been used to simulate political attitudes \citep{argyle2023out}, economic decisions \citep{aher2023using, horton2023large, korinek2023language}, psychological behaviors \citep{demszky2023using}, and social interactions \citep{ziems2024can, park2024generative}. While discrepancies between simulated and real-world behaviors remain, studies consistently show that LLM agents replicate key behavioral patterns observed in humans, making them a viable and promising proxy for human decision-makers in social simulations.

Given the key requirements for traveler agents in ABMs, LLM agents exhibit crucial capabilities that align well with this paradigm. First, their advanced natural language processing allows LLM agents to interpret both structured data and unstructured text, enabling them to incorporate diverse information sources and understand contexts. Second, LLM agents demonstrate a degree of autonomous decision-making by generating responses based on perceived information and adapting their behavior to given constraints or goals. When prompted with specific instructions or queries, they can dynamically generate responses and execute decisions accordingly. For example, in travel itinerary planning, LLM agents have demonstrated the ability to create high-quality, feasible travel plans \citep{chen2024travelagent,tang2024synergizing}. Finally, LLM agents possess role-playing ability \citep{shanahan2023role,shao2023character}, which allows them to adopt diverse personas and simulate decision-making from the perspective of different traveler profiles.

\subsection{LLM-agent-based simulation}

Leveraging LLM agents' abilities, research in physical and social sciences has integrated them into ABM frameworks for simulation \citep{gao2024large}. The goal is to use LLM agents as "guinea pigs" in multi-agent simulations to study agent interactions, emergent behaviors, and the effects of interventions \citep{grossmann2023ai}. By embedding LLM agents within these frameworks, researchers can explore how they respond to different stimuli, interact with other agents, and influence overall system dynamics. Here, we summarize and classify the current diverse applications of LLM-agent-based modeling and simulation in multi-agent interactive systems into three categories:

\begin{itemize}
    \item \textbf{Small-scale replica of laboratory experiments}: A few existing studies use LLM agents in controlled, small-scale experimental settings to observe their behavior and emergent phenomena. These experiments typically involve a limited number of agents---ranging from two to a few dozen---operating in an environment governed by fixed rules, such as game mechanics or payoff matrices. Agents in these simulations usually only make single-shot decisions in each round of interaction due to the relatively simple setting. Examples of this approach include LLM-agent-based simulations of werewolf games \citep{xu2023exploring}, transmission chains \citep{acerbi2023large}, cooperation games \citep{de2023emergent}, collaborative task-solving \citep{zhang2023exploring,chen2023agentverse}, and economic games \citep{han2023guinea,mao2023alympics,trencsenyi2025approximating}.

    \item \textbf{Neighborhood-scale or society-scale simulation of specific behaviors}: Some studies employ LLM-agent-based modeling to examine the impact of individual behaviors on system dynamics and outcomes in larger-scale systems, such as neighborhoods or small societies. In these simulations, compared to small-scale experiments, a significantly larger number of LLM agents---from hundreds to millions of agents---are involved in the simulation. However, these simulations still focus on modeling system-wide dynamics driven by a single type of agent behavior. Examples of these studies include social media platform simulations \citep{gao2023s,papachristou2024network,mou2024unveiling,tang2024gensim,yang2024oasis}, where LLM agents replicate human posting behaviors; financial markets \citep{yang2025twinmarket}, where they model investor trading patterns; public health systems \citep{williams2023epidemic,chopra2024limits}, where they simulate population's response to pandemics; international relations \citep{hua2023war}, where they act as governmental decision-makers; and national politics \citep{jiang2024casevo}, where they mimic voters' voting behavior.

    \item \textbf{Neighborhood-scale or society-scale of general human behavior}: A small body of existing research explores multi-faceted human behavior simulations using LLM agents. Unlike the previous two categories, these agents operate with greater decision-making freedom, rather than being restricted to a single type of behavior. Within comprehensive societal simulations, these frameworks capture complex agent behaviors and more complete system-wide dynamics. Notable studies in this area include: \cite{park2023generative}, which introduced a sandbox town where LLM agents interact socially and perform everyday tasks; \cite{al2024project}, which simulated society-building among multiple LLM agents in the Minecraft environment; \cite{cheng2024sociodojo} and \cite{piao2025agentsociety}, both of which developed multi-behavior agent frameworks to simulate and evaluate interactions across domains such as social networks, economics, and finance.
\end{itemize}

While LLM-agent-based simulations exist across multiple levels of social systems, for their implementation on transportation systems, characteristics of transportation systems and requirements for transportation planning applications present two unique challenges that remain unaddressed by existing studies:

\begin{itemize}
    \item \textbf{Model resolution}: Existing literature on LLM-agent-based simulations either focuses on a single behavior or attempts to model general social or economic behaviors with minimal constraints. However, travel behavior in transportation systems is inherently multi-faceted, hierarchical, and sequential. To ensure behavioral realism, the design of agents and simulation pipelines in LLM-agent-based transportation models must be expanded and refined to accurately reflect the decision-making processes of human travelers.

    \item \textbf{Model accuracy}: Most existing LLM-agent-based simulations rely on qualitative evaluation or quantitative assessments based on a single outcome, often aiming to identify emergent patterns rather than precisely replicate real-world human dynamics. However, in transportation planning, LLM-agent-based simulations must provide comprehensive, quantitative assessments of various planning scenarios. These evaluations require multi-level accuracy, including estimates of traffic flow, travel time, travel demand, transit ridership, etc. Therefore, enhancing the fidelity of LLM agents in simulating real-world traveler behavior is essential for producing dependable planning outcomes. This necessitates advanced calibration techniques to improve the accuracy of LLM-driven simulations compared to existing models.
\end{itemize}

\section{System Design} \label{sec:design}
In this section, we present the design of our LLM-agent-based transportation system modeling framework, detailing its core components, the structure of LLM agents, and the overall system workflow within the simulation, as well as outlining the key potential advantages of our proposal over existing ABM frameworks.

\subsection{Overview}

\begin{figure}[h!]
  \centering
  \includegraphics[width=0.9\textwidth]{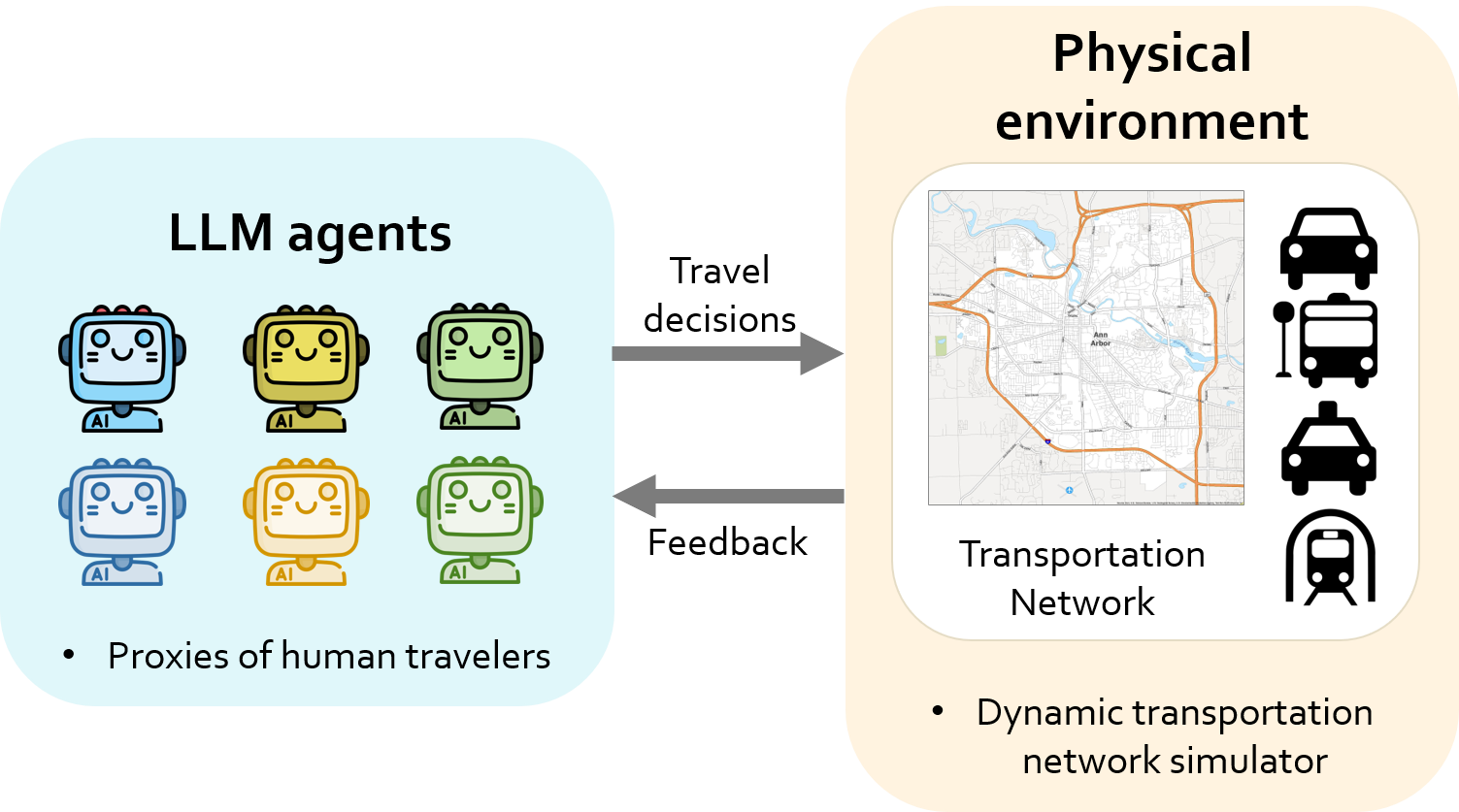}
  \caption{Overview of the LLM-agent-based modeling framework}\label{fig:system_overview_interact}
\end{figure}

Our proposed framework comprises two main components: the LLM agents and the physical environment. The agents represent distinct decision-making units within the transportation system. The physical environment, on the other hand, models infrastructure characteristics and service performance in response to evolving travel demand. In the simulation, multiple LLM agents with distinct identities, reflecting various demographic and behavioral attributes, operate concurrently and interact with the environment. This setup enables the representation of diverse population groups and their unique travel behaviors. Through these interactions, the agents and the physical environment collectively simulate complex transportation system dynamics, allowing the framework to evaluate system performance. The overall structure and agent-environment interactions within our LLM-agent-based modeling framework are illustrated in \Cref{fig:system_overview_interact}.

The core innovation of our proposed modeling framework lies in the design and implementation of LLM agents and their seamless integration into ABM systems. Before delving into the specifics of LLM agents, we first provide an overview of the physical environment, which serves as the foundation for evaluating system outcomes. The physical environment in our framework is a dynamic transportation network simulator capable of realistically representing key transportation network dynamics, including infrastructure conditions, network congestion, and multimodal transit services, while also capturing their performance responses to varying demand patterns. Furthermore, the physical environment establishes input and output requirements for key modules in LLM agent design, ensuring that the agent architecture remains compatible and functionally consistent with the broader modeling framework. The simulation environment can either be developed from scratch by the modeler or built upon existing dynamic network simulation platforms, such as SUMO \citep{krajzewicz2002sumo} or Aimsun \citep{barcelo2005dynamic}, through integration or extension.

\subsection{LLM agent}

\begin{figure}[h!]
  \centering
  \includegraphics[width=0.9\textwidth]{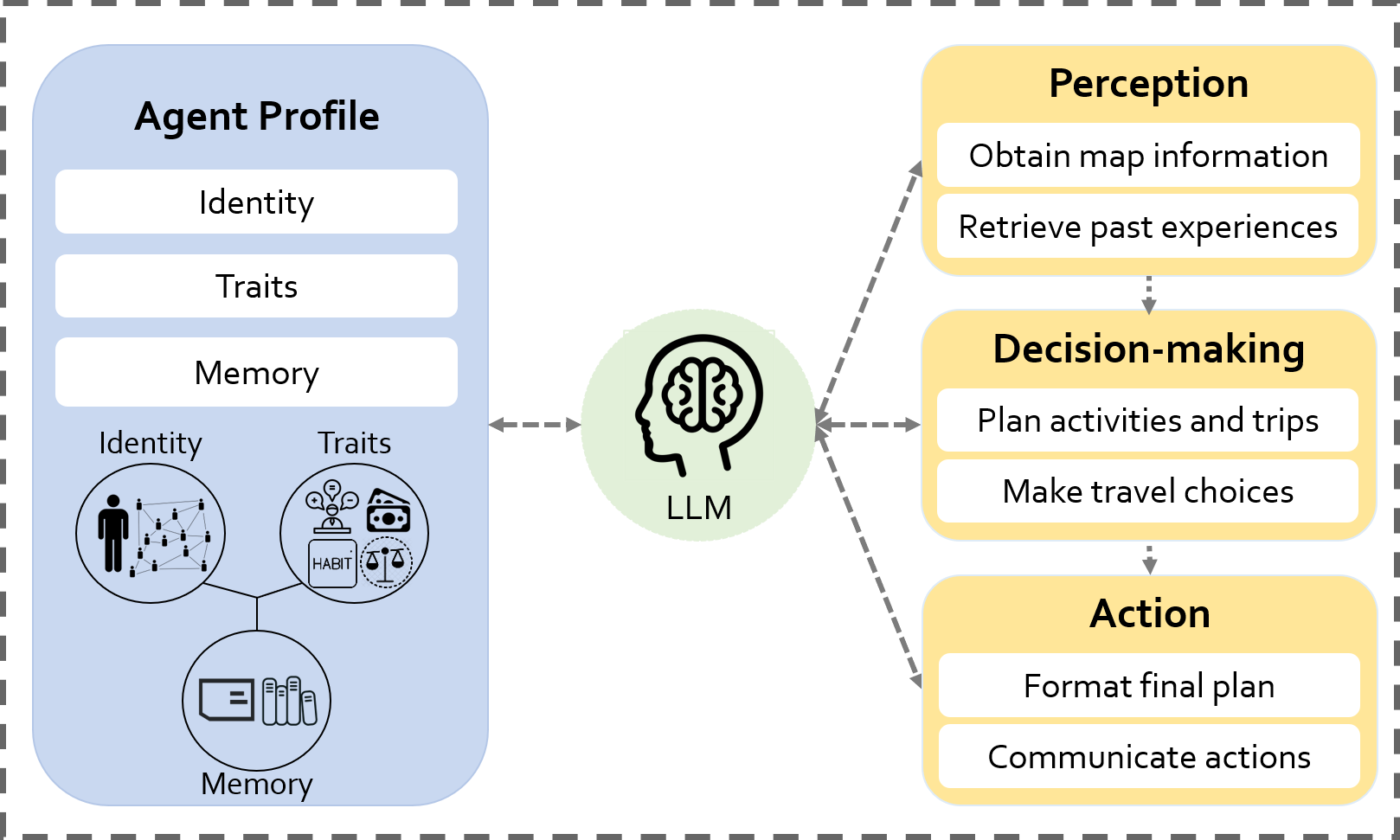}
  \caption{Design of LLM agents}\label{fig:llm_agent_design}
\end{figure}

In our proposed framework, as shown in \Cref{fig:llm_agent_design}, LLM agents are designed to emulate the behavior of human travelers. To achieve a comprehensive emulation, our LLM-based agent architecture consists of an LLM core (which is an LLM instance) and four key components that support its functionality: the profile module, the perception module, the decision-making module, and the action module. 

In the profile module, we establish a human-like profile for LLM agents that serves as the intrinsic features for shaping behavior. Human travel behavior is influenced by numerous factors, including preferences, attitudes, socio-demographic characteristics, travel purposes, habits, and past experiences, and we design the profile module to capture the impact of these factors. Specifically, each agent’s profile consists of three key components: identity, traits, and memory. The function and behavioral implications of each component are detailed below:

\begin{itemize}
    \item \textbf{Identity}: The identity module defines the core socio-demographic characteristics that shape the agent’s travel behavior. This includes information such as name, age, occupation, wage, household income, vehicle ownership, and residential location. Additionally, the identity determines obligatory travel activities. For instance, an office worker must commute to work on workdays, while a student attends school on school days. Each agent's identity is stored as natural language representations in data tables, ensuring no information loss due to aggregation while allowing seamless integration with the LLM core for contextual understanding and role-playing in behavioral simulation.
    
    \item \textbf{Traits}: Beyond identity, an agent’s profile incorporates attitudes, habits, decision-making tendencies, economic preferences, and trade-offs, as these factors significantly influence travel behavior \citep{mokhtarian2024pursuing,ben2015response,aarts1996habit}. Attitudes reflect an agent’s perceptions of travel-related aspects, such as preference for travel enjoyment or environmental consciousness. Habits capture routine travel patterns, such as car dependency. Decision-making traits encapsulate aspects of bounded rationality, including cognitive inertia, degree of loss aversion, and risk preference. Finally, economic preferences and trade-offs characterize the agent’s economic behavior, such as the value of travel time savings (VTTS). Like identity, these trait attributes are stored in natural language format, enabling efficient retrieval and integration during simulation.
    
    \item \textbf{Memory}: In addition to the attributes, agents dynamically learn and adjust based on past travel experiences. For instance, frequent delays on a particular route may prompt a traveler to adjust departure times or switch modes in future trips. To replicate this learning process, we implement a memory module that tracks and stores agents' past experiences and reflections. The memory system consists of two layers: short-term memory, which records detailed recent travel experiences, and long-term memory, which retains high-level summaries and insights that persist over time. All stored memories are structured in natural language format, following a predefined data structure to facilitate retrieval and update during decision-making.
\end{itemize}

In transportation systems, travelers perceive information from the system and make travel-related decisions based on their evaluation of alternatives, which depends on their intrinsic attributes and the perceived attributes of the alternatives. Other than making decisions, travelers must also act upon them and carry them out through interaction with the physical environment. Therefore, beyond maintaining a comprehensive agent profile, an agent's ability to process environmental information, make informed decisions, and execute actions is essential. To achieve this, we introduce three vital functional modules: the perception module, the decision-making module, and the action module. The LLM core serves as the central component, playing a critical role in all functional modules. The core functionalities of these modules are described below:
\begin{itemize}
    \item \textbf{Perception}: Travelers frequently acquire real-time information about the physical environment---such as points of interest or candidate routes---before making travel decisions, and they also receive feedback on the outcomes of past actions. To reflect this, we equip agents with corresponding perceptual capabilities, allowing them to query and extract relevant information from the physical environment. This data is initially retrieved in a structured format (e.g., numerical vectors). Next, the LLM core processes this structured information, translates it into natural language, and incorporates it into subsequent decision-making processes. The perception module can also interact with the memory system, retrieve past travel experiences, and organize them to prepare for decision-making.
    
    \item \textbf{Decision-making}: Human travel behavior contains a series of multi-faceted decisions. To simulate this using LLM agents, we combine prompt engineering with strategic querying in the decision-making module. For each travel decision, we retrieve the agent’s identity attributes, behavioral traits, and past experiences from its profile and integrate them with perceived environmental data into a natural language prompt. This prompt instructs the LLM core to simulate the agent’s travel behavior, generate a structured response, and adhere to predefined output formats. The LLM’s capabilities in contextual understanding, planning, evaluation, and role-playing allow it to produce nuanced, dynamic responses that reflect adaptive reasoning and realistic decision-making strategies.

    \item \textbf{Action}: While LLMs can simulate travel behavior, their outputs are in natural language, whereas dynamic transportation network simulators used in our physical environment require structured inputs like time-dependent origin-destination (O-D) trip matrices. To bridge this gap, the action module translates LLM-generated text into simulator-compatible formats. The action module parses LLM-generated text, extracts key travel parameters, including departure time, destination, mode choice, etc., and maps them into standardized data structures that satisfy the simulator's input. This ensures that the agents' behaviors are accurately reflected in the physical environment and the simulation.
\end{itemize}

\subsection{System workflow}

\begin{figure}[h!]
  \centering
  \includegraphics[width=0.9\textwidth]{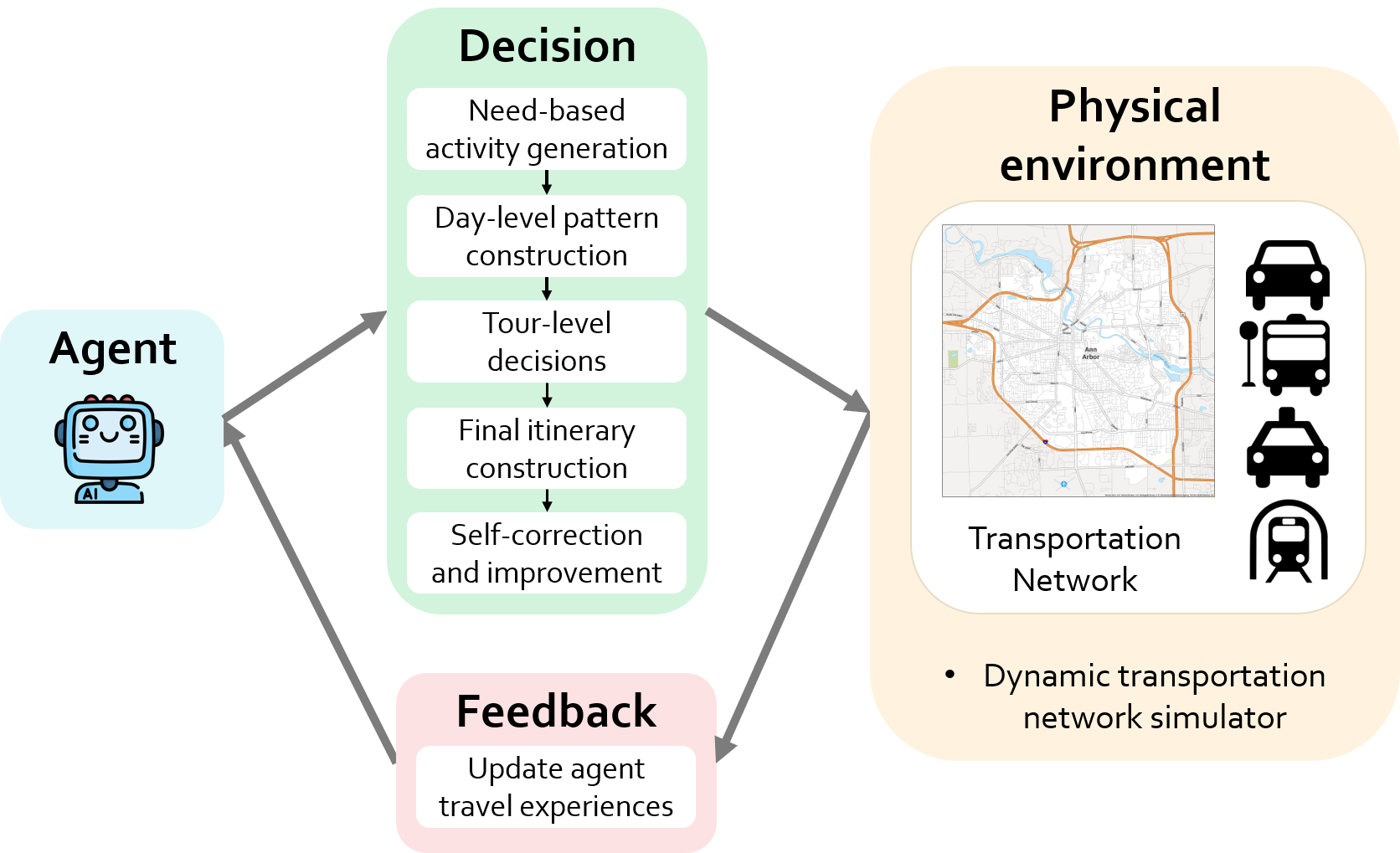}
  \caption{Workflow of the simulation system}\label{fig:system_workflow}
\end{figure}

In designing the system workflow of our LLM-agent-based model, we integrate the principles of activity-based travel demand models \citep{castiglione2015activity} to enhance behavioral realism in both agent decision-making and agent-environment interactions. An overview of the system workflow is presented in \Cref{fig:system_workflow}.

In \Cref{fig:system_workflow}, the behavior of agents and their interaction with the transportation infrastructure follows the general workflow of activity-based models. The proposed framework primarily focuses on day-level decision-making and system dynamics; however, it can be extended to incorporate long-term decisions (e.g., residential location choices) and medium-term choices (e.g., transit memberships and vehicle ownership) by adding supplementary functions within the same pipeline. The decision-making and interaction flow of agents consists of the following key steps:
\begin{itemize}
    \item \textbf{Need-based activity generation}: Travel demand is largely derived from individuals’ daily activities, which are shaped by socio-demographics, lifestyle preferences, and past experiences \citep{arentze2009need, liu2020day}. To reflect this characteristic of travel behavior, in the initial stage of decision-making, LLM agents assess household and traveler characteristics and generate a comprehensive set of activities, encompassing both mandatory obligations (e.g., commuting to work or school) and discretionary choices (e.g., dining out or shopping). Agents reference related profiles and draw upon relevant experiences to formulate daily activities that align with individual and household-level needs.

    \item \textbf{Day-level pattern construction}: In activity-based models, structuring daily travel patterns is a key step between activity generation and trip itinerary formulation \citep{arentze2000albatross, castiglione2015activity}. It defines household- and individual-level travel attributes such as the number of tours, activity sequences, and the distribution of travel responsibilities \citep{gali2008activitysim, auld2012activity}. To capture this process, LLM agents integrate previously generated activities with past travel experiences to construct coherent, realistic daily travel patterns. Through contextual reasoning and learned heuristics, agents organize household activities into structured tours, carefully selecting participants for each tour, activity durations, and sequencing constraints.

    \item \textbf{Tour-level decisions}: Once day-level travel patterns are established, the next step involves specifying tour attributes \citep{gali2008activitysim,castiglione2015activity}. At this stage, the LLM agents synthesize tour-level details with present-day constraints, optimizing schedules to balance efficiency, feasibility, and personal preferences. By continuously referencing spatial knowledge, habitual patterns, and day-level constraints, agents formulate adaptive and individualized tour schedules, which guide the subsequent final itinerary generation.

    \item \textbf{Final itinerary construction}: The final phase in activity-based modeling involves completing detailed travel itineraries. In our framework, agents finalize their tour schedules by specifying key trip attributes such as departure and arrival times, travel modes, and route choices. They incorporate relevant memories and contextual cues in constructing the final itineraries, allowing agents to make informed decisions based on prior experiences and current travel conditions. This step produces a coherent and realistic trip chain for each traveler, which can then be converted to O-D matrices in the action module.

    \item \textbf{Self-correction and improvement}: LLM agents may occasionally produce unrealistic or infeasible travel plans. To improve reliability, we implement a self-correction mechanism in which agents undergo a post-processing phase to review and correct inconsistencies and impracticalities in their schedules, leveraging LLMs’ self-refinement abilities \citep{kamoi2024can}. Our previous work \citep{liu2025LLMABM} shows that this process significantly improves the accuracy and coherence of LLM-generated travel plans.

    \item \textbf{Feedback}: After itinerary generation, agents' travel plans are converted into structured inputs for simulation in the physical environment. The resulting travel outcomes, such as congestion levels, delays, and arrival times, are fed back to the agents. This feedback is then processed into the agents' memory systems, enabling adaptive learning and continuous refinement of future travel choices. By incorporating system feedback, LLM agents progressively enhance their decision-making strategies, ensuring that subsequent travel behaviors reflect evolving conditions and learned experiences.
\end{itemize}

\subsection{Key potential advantages}

We argue that our approach of integrating LLMs into transportation ABMs offers more than just compatibility---LLM-based agents can enhance existing frameworks and help address key limitations of traditional models. Below, we highlight several key potential advantages of our approach.

\subsubsection{Relaxing behavioral assumptions}
LLM agents have the potential to relax the assumptions embedded in behavior models applied in existing ABMs. As discussed in \Cref{subsec:LR_existing_ABM}, traditional ABMs rely on econometrics-based models and rule-based frameworks, both of which impose assumptions about traveler behavior and have limitations in capturing the nuanced behavior. LLM agents offer an alternative that can relax some assumptions and incorporate a broader range of behavioral factors into travel demand modeling. Trained on vast corpora of human-generated text and record, LLMs are exposed to an extensive dataset that inherently reflects human thought patterns, preferences, and social behaviors. Therefore, in the training process, LLMs could learn from these rich patterns in the training data and approximate human language and behavior as a result. This characteristic makes them well-suited to capturing the complex and multi-factorial behavior exhibited by human travelers. Empirical studies further support this potential, with research showing that LLMs exhibit economic decision-making patterns similar to those observed in humans \citep{horton2023large, coletta2024llm}, as well as human-like cognition \citep{strachan2024testing} and social behavior \citep{leng2023llm, zhang2023exploring, xie2024can}. These findings highlight the potential of LLM agents to enhance behavioral realism in ABMs by relaxing assumptions and better capturing the complexity of human travel behavior.

\subsubsection{Better leveraging existing data sources}
LLM agents offer the potential for more efficient and expanded utilization of existing data sources in the calibration process compared to conventional approaches. In existing ABMs, behavioral models require extensive data collection in terms of contexts, which can be a challenge, particularly for agencies or communities with limited resources. LLM agents can potentially mitigate this challenge for ABMs. First, LLM agents may leverage extensive pre-training on diverse human-generated datasets, encompassing a broad range of knowledge domains and behavioral contexts. This pre-training equips LLMs with a rich baseline of generalizable insights into human decision-making, thereby reducing the need for data-intensive calibration from scratch. Moreover, LLM agents can incorporate heterogeneous data sources, including unstructured data such as textual content, images, and other multimodal inputs. This capability allows model developers to further optimize training by leveraging existing data resources. Finally, the few-shot learning capabilities of LLMs enable them to generalize behaviors across diverse settings with minimal additional data. Consequently, LLM agents have the potential to broaden the accessibility of model training, enabling agencies and communities with resource constraints to implement agent-based simulations.

\subsubsection{Better supporting alternative evaluation}

The LLM-agent-based framework also facilitates the evaluation of alternative transportation scenarios and policies more efficiently compared to conventional methods. Forecasting travel demand and assessing system performance under new plans or policies, especially those involving emerging technologies like connected and automated vehicles and on-demand public transit, often poses challenges. Established ABMs may require extensive modifications in both agent configurations and supporting datasets to adapt to these new scenarios. In contrast, LLM agents provide significant flexibility, allowing agent behaviors to be customized through natural language prompts rather than complex logic or code specific to each scenario. This capability reduces the time and expertise required for setting up sophisticated simulations. Additionally, the few-shot and zero-shot prompting mechanism of LLMs enables generalization to alternative situations with minimal data. This flexibility reduces the necessity of explicitly programming detailed behavioral rules and processes for each scenario. Several studies have demonstrated the effectiveness of LLM-agent-based modeling for evaluating complex policies and scenarios. For example, \cite{han2023guinea} explored communication policies and market collusion; \cite{hua2023war} simulated historical international events; \cite{li2024econagent} assessed COVID-19’s impact on U.S. unemployment; and \cite{chopra2024limits} modeled public health outcomes of containment policies in New York City. In each case, policies and scenarios were provided as natural language inputs, significantly simplifying simulation setup.

In addition, an LLM-agent-based framework also opens up opportunities for interactive planning interfaces, where transportation planners and policymakers can iteratively engage with the model throughout the planning process. In this interface, planners can query the model in natural language, propose policy interventions, or adjust assumptions dynamically, with the LLM agents responding to the planners adaptively based on their needs and requirements. This two-way interaction supports a more adaptive and collaborative approach to transportation planning, enabling planners to evaluate ``what-if" scenarios on the fly, receive feedback for agent behaviors or system-level outcomes that are customized to the planner’s needs, and potentially refine policies through conversational feedback loops.

\subsubsection{Supporting learning of agents}
An essential feature of ABMs in transportation systems is the ability of agents to learn and adjust their travel behavior over time. Beyond simulating the learning process of human travelers, adaptive agents play a crucial role in facilitating system convergence toward equilibrium and enabling consistent and fair assessments of alternative scenarios. However, efficiently modeling learning and behavioral adjustment remains a challenge in existing ABM frameworks. Traditional methods, such as including random permutations and co-evolutionary algorithms, could suffer from limited efficiency or generalization. LLM agents present a promising alternative for modeling learning and behavioral adaptation, as they possess advanced contextual understanding, reasoning abilities, and intelligence that allow them to infer system patterns and make strategic adjustments to their travel decisions. Compared to traditional approaches, LLM agents offer both high adjustment efficiency and strong generalization capabilities, making them more effective in simulating adaptive human-like decision-making. These advantages are supported by existing literature. For example, \cite{patwary2024bridging} studied agent planning of electric vehicle charging in Greater Montreal Island, Canada, and found that LLM-based planning achieved a higher success rate and produced more reliable behavioral adjustments than conventional methods.

\section{Key Validations on Behavioral Alignment} \label{sec:prop}

In designing LLM agents for transportation system modeling, a crucial task is to have them closely replicate human travel behaviors. As agent-environment interactions in transportation systems are inherently iterative, agents must not only make travel decisions but also incorporate system feedback to refine their strategies over time. Thus, agent behavior should align with that of human travelers along three critical dimensions: First, LLM agents should exhibit a similar distribution of travel-related activity types, frequencies, and timings as human travelers while controlling for individual identities; Second, their travel choices should be consistent with those made by human travelers under comparable conditions; Third, LLM agents must demonstrate a human-like ability to learn from past experiences and adjust their decisions accordingly.

While LLMs have been trained on vast amounts of human-generated data, there is no guarantee that their behavior will naturally resemble that of humans. Even with role prompting \citep{liu2023pre}, where socio-demographic details of the decision-maker are provided to enhance role-playing, studies from social science and psychology \citep{hagendorff2023human,goli2023can,chen2023emergence,tjuatja2024llms,park2024diminished,taubenfeld2024systematic,wang2024large,fan2024can} suggest that merely prompting LLMs with task descriptions, demographics, and contextual details may not be sufficient to generate human-like behaviors. Consequently, additional tuning of LLM agents is required to better align their travel choices with human travelers. In this section, we review existing research on this topic and assess the feasibility and progress of designing and training LLM agents to align well with humans across the three key dimensions outlined above.

\subsection{Activity generation and scheduling}
Existing research on LLM-based activity generation focuses on tuning and conditioning LLM agents to produce travel diaries that resemble those of human travelers, as these diaries capture related activities and serve as a useful proxy for understanding activity generation and scheduling. Encouragingly, these studies indicate that with targeted prompting, conditioning, and fine-tuning, LLM agents can effectively simulate activities across a wide range of contexts. 

Some research integrates in-context learning with behavior theory to enhance LLM agents’ ability to emulate human activity patterns. In \cite{wang2024large}, an agentic framework combining memory and motivation was employed to prompt LLM agents in generating human-like mobility trajectories. The framework first learned from real-world trajectory data and socio-demographic information, allowing LLM agents to establish habitual travel patterns within a population. Each agent was then assigned a pattern that best aligned with its past activities. During travel diary generation, the agent identified motivations and travel needs based on socio-demographic attributes, the date, and its habitual pattern. These components were then incorporated into a structured prompt to generate a new daily travel diary. Evaluations on a large-scale mobility dataset revealed that this tuning approach produced activity patterns more aligned with human behavior than existing theory-based or machine-learning-based approaches. Building on this approach, \cite{li2024more} introduced learning-based and persona-driven LLM agents to further enhance emulation. Their agentic framework extended the behavioral foundation by not only analyzing habitual travel patterns but also incorporating personalized factors such as preferences and responsibilities. These learned personas were later embedded into prompts for travel diary generation. Evaluation on real-world household travel survey data showed that this method significantly outperformed deep-learning-based techniques in generating human-like activities.

Even in the absence of large-scale individual travel diaries, LLM agents can still generate activity patterns similar to human travelers. \cite{liu2024human} demonstrated that with few-shot prompting and high-level contextual guidance, LLM agents could generate reasonable travel diaries that align with real-world mobility patterns. Their prompt design incorporated the agent's socio-demographic profile; economic and mobility statistics at a granular level; guidelines incorporating human knowledge; and few-shot examples of travel trajectories. Using the NHTS and the SCAG travel survey data, they showed that few-shot prompting enabled LLM agents to generate activity types and locations that closely resembled those observed in human travelers.

Beyond few-shot prompting, fine-tuning techniques offer another promising avenue for improving LLM-based travel demand generation. \cite{zhang2024agentic} combined few-shot prompting with fine-tuning by training an open-source LLM model on a small set of travel diary data. Their findings indicated that the fine-tuned model could produce activity distributions similar to the human-generated ground truth, though LLMs tended to generate fewer trips per day than human travelers.

Since travel behavior is largely a derived demand driven by human needs \citep{mokhtarian2001derived}, generating accurate activity patterns also hinges on an LLM agent’s ability to infer and respond to human needs. Encouragingly, research suggests that LLM agents also exhibit promising capabilities in this area. For instance, \cite{wang2024simulating} developed an LLM agent that integrated human desire modeling with a qualitative value system to simulate daily activity generation. Their evaluation confirmed that LLM agents could generate activities that fulfilled emerging needs, exhibiting variability and adaptability akin to human behavior.

\subsection{Travel choices}
On choice modeling, existing research has demonstrated that LLM agents, when integrating behavioral theories and data mining techniques, can exhibit travel choice behaviors similar to those of human travelers.

\begin{figure}[h!]
  \centering
  \includegraphics[width=1\textwidth]{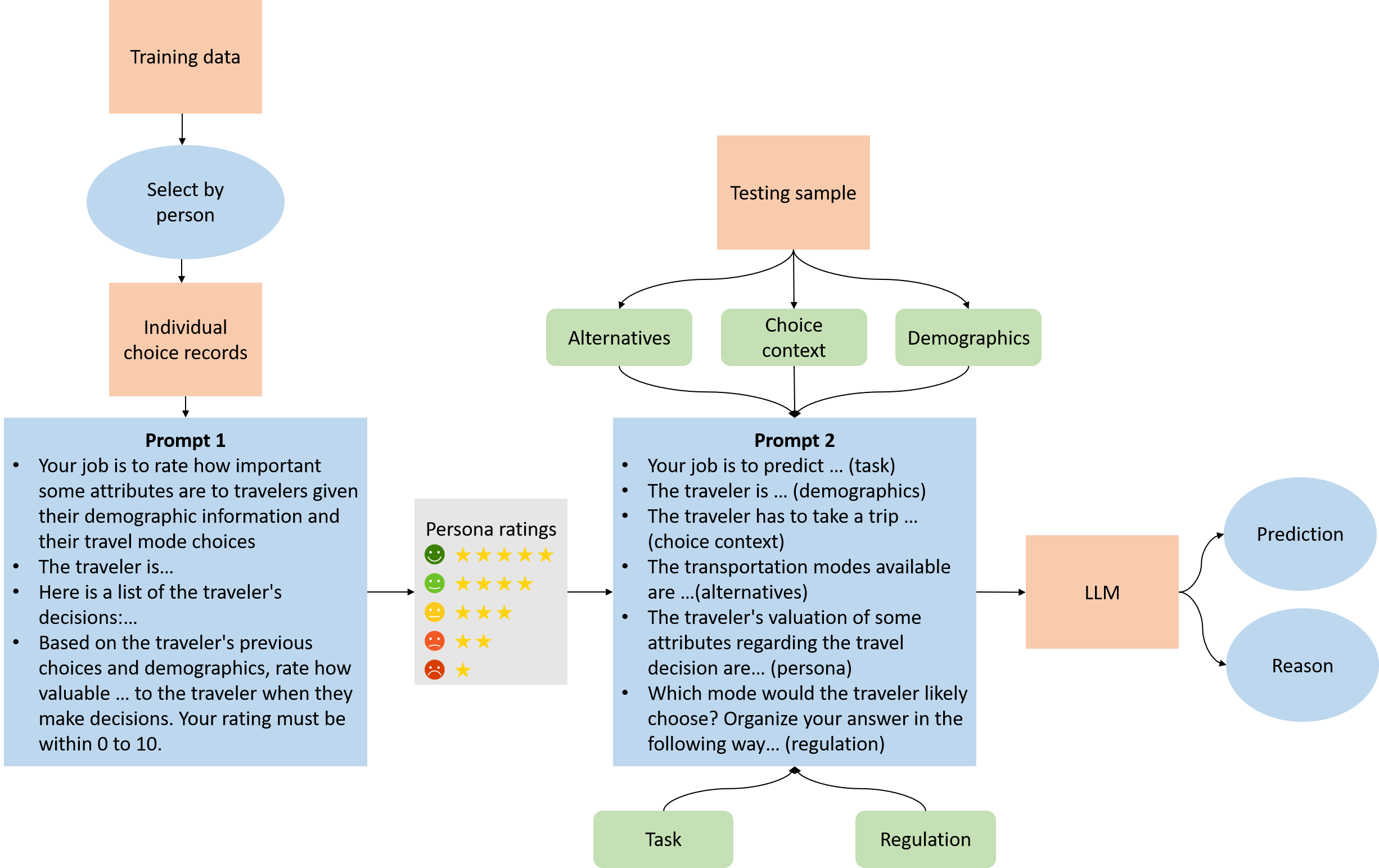}
  \caption{The LLM agent mode choice prediction framework in \cite{liu2024can}}\label{fig:persona_loading_framework}
\end{figure}

In our previous study on mode choice \citep{liu2024can}, we found that role prompting based solely on socio-demographic information and few-shot learning was insufficient to fully align LLM-generated mode choices with those of human travelers. To address this limitation, we proposed a persona discovery and loading framework (illustrated in \Cref{fig:persona_loading_framework}) to further tune LLM agents on mode choice emulation. The framework first leveraged LLMs’ reasoning capabilities to identify latent persona characteristics in mode choice behavior. This was achieved by prompting the LLM agent to summarize the revealed economic trade-offs based on the observed mode choices of different individuals, effectively clustering them into distinct personas. Subsequently, new observations were matched to the most suitable latent persona, akin to a latent class discrete choice model. The LLM agent then role-played using both the traveler’s socio-demographic profile and the assigned persona to make mode choice decisions. Empirical validation using real-world data indicated that this framework significantly enhanced the behavioral alignment between LLMs and human travelers.

LLM agents also demonstrate strong predictive capabilities in destination choice modeling. Using a few-shot learning approach, \cite{wang2023would} investigated whether LLMs could predict a traveler’s next destination based on their current location and past activities. Their framework integrated historical travel data and domain-specific knowledge, enabling the LLM to extract patterns and infer likely destination choices. The LLM agent was prompted to role-play as the traveler and make destination predictions based on past trajectories, current location, and contextual guidance. Empirical validation on real-world mobility data confirmed that LLMs generated destination choices comparable to those of human travelers. Similarly, \cite{beneduce2024large} employed a few-shot learning framework and reported comparable findings. Moreover, they demonstrated that zero-shot learning approaches can also produce high-quality predictions. Subsequent research refined these methods by incorporating memory structures, geospatial knowledge, and data mining techniques to enhance LLMs’ behavioral realism and role-playing abilities. For example, \cite{qin2025lingotrip} integrated geospatial domain knowledge within prompts to enhance CoT reasoning for zero-shot next-location prediction, finding that LLMs equipped with spatial knowledge effectively predicted travelers' destination choices; \cite{feng2024agentmove} combined memory-based tracking with an environment knowledge generator in an agentic framework, ensuring that LLM agents produced more realistic destination choices; \cite{chen2025toward} introduced an entropy-based choice set generation method, allowing LLM agents to construct more accurate and diverse destination choice sets.

Beyond few-shot prompting, fine-tuning techniques are also explored to improve the accuracy of LLM agents in destination choice modeling. \cite{gong2025mobility} developed a fine-tuned LLM agent that integrated memory and preference-based embeddings for destination prediction. In their framework, geospatial data and travel purposes were vectorized and fed into a fine-tuning model. The LLM agent was optimized using low-rank approximation and GPS travel data, allowing it to learn personalized travel preferences. By incorporating preference pairs that encoded individual traveler characteristics, the fine-tuned LLM model significantly outperformed state-of-the-art machine learning models in predicting destination choices across multiple GPS tracking datasets. Similarly, \cite{liu2024nextlocllm} employed fine-tuning with LLM-generated embeddings to enhance the accuracy of destination choice prediction.

Additional evidence of LLMs’ behavioral alignment with human travelers is observed in other travel choice contexts. For example, \cite{chen2024delayptc} explored how LLMs can simulate traveler decision-making during train delays using a few-shot learning framework. By incorporating delay logs, contextual features, domain knowledge, and CoT prompting, the LLM was tasked with predicting traveler choices during delay events. These features enabled the model to identify key patterns and improve behavioral prediction. Validation with real-world data showed that delay-log-tuned LLMs accurately simulated waiting decisions.

\subsection{Learning and adjustment}

Existing research in transportation and game theory suggests that LLM agents can be conditioned to learn and adjust their behavior based on interactions with environments, exhibiting human-like and reasonable adjustments in decision-making patterns.

\begin{figure}[h!]
  \centering
  \includegraphics[width=0.9\textwidth]{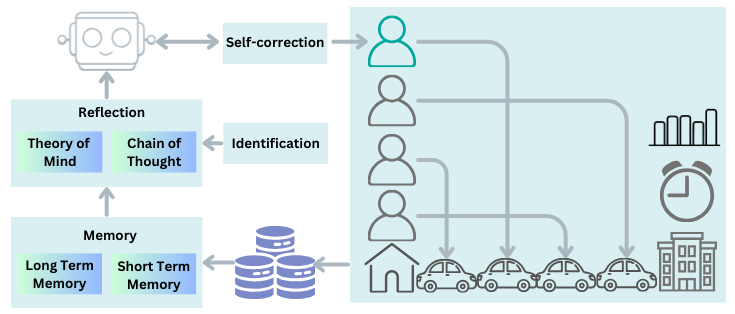}
  \caption{The agent and game setting in \cite{liu2025LLMABM}}\label{fig:dtd_agent_framework}
\end{figure}

In transportation research, researchers have built LLM agents and tested their learning and adjustment behavior in day-to-day route choice and departure time choice contexts. \cite{wang2024ai} designed an LLM agent that incorporates exponential smoothing memory structures and prompt-based exploration-exploitation balance strategies to model route choice behavior in a day-to-day environment of multiple agents. Their experiments in single O-D and network settings showed that agents learned from past interactions and reached near-equilibrium after initial exploration. Furthermore, in the single O-D case, agents' route-switching behavior aligned with human patterns. In the day-to-day departure time choice setting, \cite{liu2025LLMABM} developed an LLM agent equipped with a memory system with a self-reflection feature, incorporating theory-of-mind and behavioral inertia as part of its decision-making traits (as illustrated in \Cref{fig:dtd_agent_framework}). In a bottleneck scenario, they observed that after a brief exploratory phase, LLM agents quickly adjusted their departure times, converging to near equilibrium. The agents demonstrated the ability to learn from past experiences, refine their strategies, and stabilize their choices in a manner consistent with equilibration processes.

While research on LLM agents' learning and adaptation in transportation settings remains limited, extensive studies in game theory, particularly in repeated game settings, provide strong evidence of their capability for strategic learning and adjustment. Although vanilla LLMs may struggle with learning and adaptation in multi-agent game settings \citep{fan2024can}, empirical studies have indicated that LLM agents, equipped with human-like perception and decision-making modules, can improve their strategies over repeated interactions and that this learning ability generalizes across different game environments \citep{akata2023playing,duan2024gtbench,huang2024far}. Additionally, LLM agents can be constructed to imitate human strategic behavior in competitive settings. For instance, \cite{mao2023alympics} demonstrated that LLM agents, when prompted with persona-based role descriptions and equipped with memory structures, exhibit game-playing styles closely resembling those of human players in an auction game. Similarly, \cite{trencsenyi2025approximating} implemented a guided CoT mechanism for LLM agent decision-making, revealing that LLM-generated strategies in a two-person beauty contest game closely mirror human behavior.

\section{System Demonstration} \label{sec:demonstration}

To illustrate the feasibility and key properties of the proposed LLM-agent-based modeling framework, this section presents a demonstrative prototype consisting of 10 LLM agents operating within a transportation network of a miniature metro area. This setup is designed as a controlled, small-scale environment to evaluate whether the agents are capable of generating coherent, context-aware travel behavior through interaction with the physical environment. Rather than aiming for a comprehensive large-scale simulation, this demonstration serves as a proof-of-concept, highlighting the agents' ability to:
\begin{enumerate}
    \item Autonomously generate travel demand based on their individual profiles;
    \item Format outputs and interact effectively with the infrastructure environment;
    \item Learn from feedback on travel outcomes and adjust travel behavior accordingly.
\end{enumerate}

By isolating key elements of agent behavior in a manageable setting in this proof-of-concept, we aim to qualitatively assess the effectiveness of the proposed framework. This framework serves as an initial step toward validating the adaptability of LLM agents in ABMs of transportation systems.

\subsection{Simulation setup}

\subsubsection{Transportation network}

\begin{figure}[h!]
  \centering
  \includegraphics[width=1\textwidth]{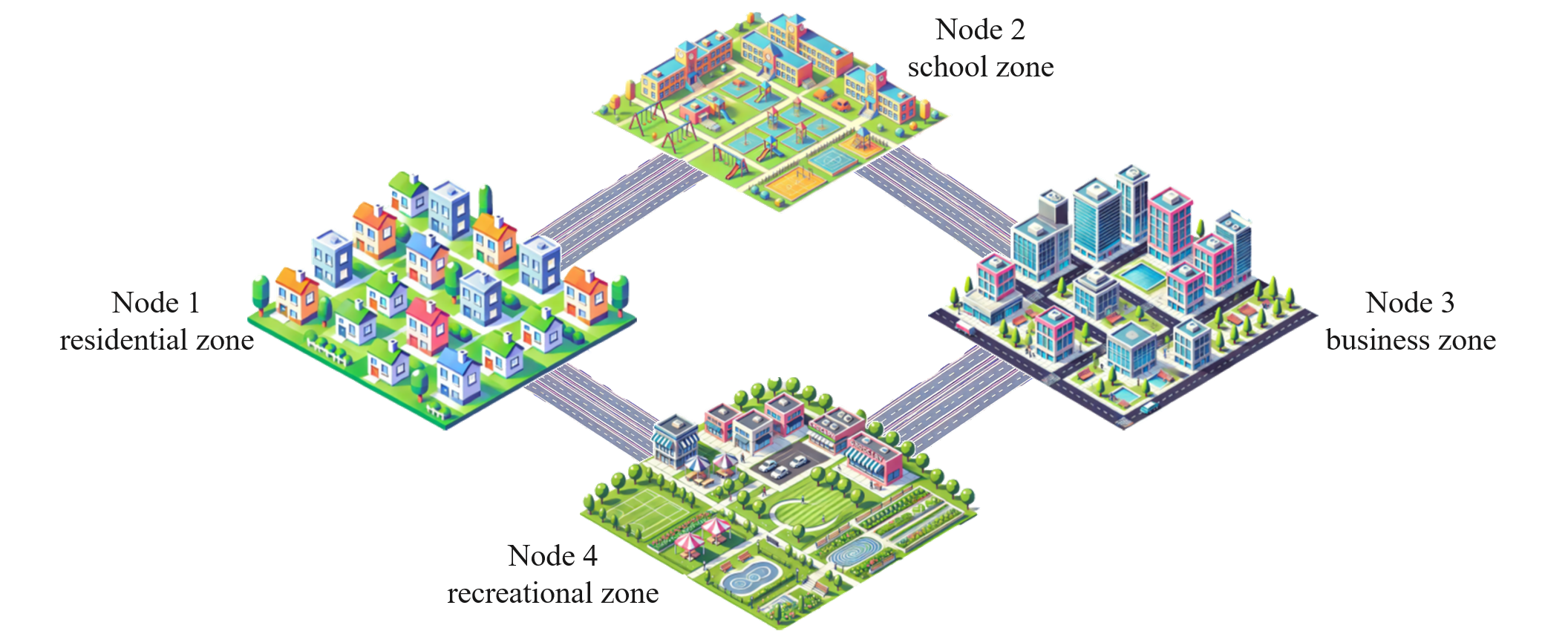}
  \caption{Illustration of the transportation network}\label{fig:example_network}
\end{figure}

In the demonstrative example, we use a small transportation network consisting of four zones to represent a miniature metro area and its transportation infrastructure. An illustration of the network is shown in \Cref{fig:example_network}.

The transportation network connects four zones, each representing a distinct function:
\begin{itemize}
    \item \textbf{Zone 1 (Residential Zone)} – Home location of all agents;
    \item \textbf{Zone 2 (School Zone)} – Location of schools in the area;
    \item \textbf{Zone 3 (Business Zone)} – Workplace for agents with in-office jobs;
    \item \textbf{Zone 4 (Recreational Zone)} – Destination for recreational and some maintenance activities such as dining out, grocery shopping, or watching a movie.
\end{itemize}

All links in the network support car travel, and their characteristics are detailed in \Cref{tab:link_info}.

\begin{table}[h!]
\centering
\renewcommand{\arraystretch}{1.2} 
\begin{tabular}{@{}l c c c c@{}}
\hline
Link ID & Origin zone & Destination zone & Free-flow travel time (mins) & Capacity (veh/h) \\ 
\hline
1 & 1 & 2 & 20 & 80\\
2 & 1 & 4 & 20 & 80\\
3 & 2 & 3 & 40 & 80\\
4 & 4 & 3 & 40 & 80\\
5 & 2 & 1 & 20 & 80\\
6 & 4 & 1 & 20 & 80\\
7 & 3 & 2 & 40 & 80\\
8 & 3 & 4 & 40 & 80\\
\hline
\end{tabular}
\caption{Link setup for the transportation network}
\label{tab:link_info}
\end{table}

\subsubsection{Agent profile}

\begin{figure}[h!]
  \centering
  \includegraphics[width=0.75\textwidth]{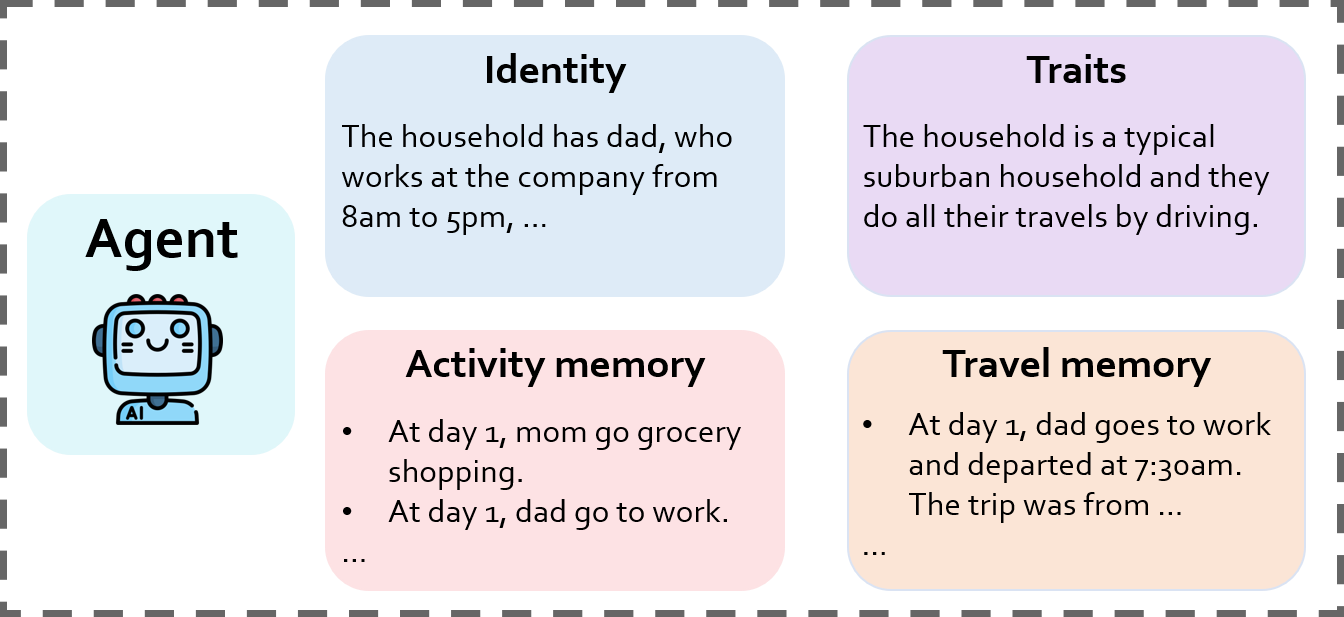}
  \caption{Illustration of LLM agent used in the demonstrative example}\label{fig:example_agent_design}
\end{figure}

We use a total of 10 LLM agents driven by GPT-4o, numbered from 1 to 10, in the demonstrative example. Each agent represents a decision-making unit, which in this case is a household. On agent profiles, we assign agents similar profiles that generate the same mandatory activities, thereby inducing overlapping travel demand and congestion. This setting creates a meaningful environment for evaluating the role of agent learning and adaptive behavior in the face of congestion. At the same time, we introduce variation in the traits of some agents to assess their contextual understanding of profiles and their ability to generate non-compulsory activities. All agents share the same identity structure and memory system. The general setup of the agent profile is illustrated in \Cref{fig:example_agent_design}.

The shared identity of all agents is:
\begin{quote}
    ``The household has dad, who works at the company from 8 am to 5 pm, mom, a self-employed worker who work from home from 10:30 am to 3:30 pm, and kid, who attends high school from 8 am to 4 pm."
\end{quote}

Each agent is equipped with two parallel memory systems to support different steps in travel demand generation and decision-making:
\begin{itemize}
    \item \textbf{Activity memory}: This memory stores all past activities undertaken by the household, regardless of whether travel is involved. Each activity is recorded as a natural language entry that includes the activity's date and detailed contextual information.
    \item \textbf{Travel memory}: This memory stores the outcomes of all past trips. Each trip is recorded in natural language, including information such as the date, participants, purpose, origin and destination zones, route taken, departure and expected arrival times, actual arrival time, and total travel time.
\end{itemize}

\begin{figure}[h!]
  \centering
  \includegraphics[width=0.7\textwidth]{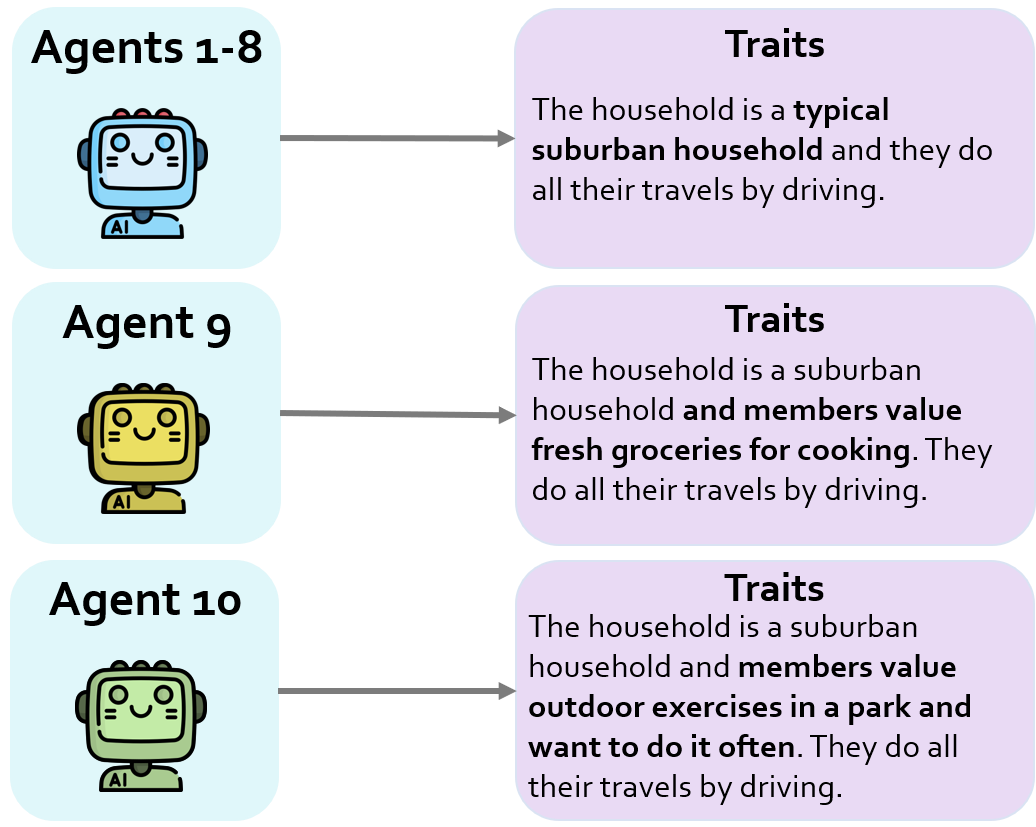}
  \caption{Variations in agent traits used in the demonstrative example}\label{fig:example_agent_vary}
\end{figure}

While all agents share the same mandatory activities---such as commuting to work and school---variation is introduced in the agent traits to assess the LLM agents' capacity to generate non-mandatory activities based on individual preferences and needs. \Cref{fig:example_agent_vary} illustrates the variation in agent traits used in the experiment.

For agents 1-8, we give them the base trait:
\begin{quote}
    ``The household is a typical suburban household and they do all their travels by driving."
\end{quote}

Agents 9 and 10 are given modified traits to reflect specific preferences related to non-mandatory activities. Specifically, agent 9 is configured to prioritize grocery shopping (a maintenance activity):
\begin{quote}
    ``The household is a suburban household and members value fresh groceries for cooking. They do all their travels by driving."
\end{quote}

Agent 10, on the other hand, is designed to prioritize outdoor exercise (a discretionary activity):
\begin{quote}
    ``The household is a suburban household and members value outdoor exercises in a park and want to do it often. They do all their travels by driving."
\end{quote}

\subsubsection{Simulation pipeline}

We simulate the behavior of the LLM agents and the dynamics of the transportation network over 21 days, where each day corresponds to one iteration. Guided by the simulation pipeline proposed in \Cref{fig:system_workflow}, the simulation process in each iteration proceeds as follows:

\begin{enumerate}
    \item \textbf{Activity generation:} All LLM agents concurrently generate their daily activities based on their profiles and past activity records. Relevant past activities are retrieved from each agent's activity memory.
    
    \item \textbf{Pattern construction and tour formulation:} Conditional on the generated activities, agents assign activities to household members, identify which activities require travel, and formulate tours accordingly. This step leverages the agent’s understanding of household roles and scheduling constraints in the generation process.
    
    \item \textbf{Travel plan generation:} Based on the formulated tours, agents generate a complete travel plan, specifying trip origin, destination, departure time, and route selection. This decision-making process incorporates the agent’s profile and draws on past travel experiences retrieved from its travel memory.
    
    \item \textbf{Self-correction and formatting:} Agents review their generated travel plans to identify and correct potential errors or inconsistencies. They then convert the finalized plans into a structured format suitable for input to the physical environment.
    
    \item \textbf{Physical environment:} The structured travel plans of all agents are merged and fed into the physical environment (a dynamic traffic assignment simulator). In this example, we use DTALite \citep{zhou2014dtalite} to simulate traffic dynamics and travel outcomes.
    
    \item \textbf{Feedback and memory update:} After the simulation, travel outcomes are returned to the respective agents. Each agent updates its travel memory based on the outcomes, enabling learning and adjustment in future iterations.
\end{enumerate}

\subsection{Simulation results and discussion}
\subsubsection{Activity generation}

In our setting, each agent is expected to generate two mandatory activities that require travel each day: the dad commuting to work and the kid going to school. To evaluate whether LLM agents can properly handle mandatory activity generation, we examine whether agents consistently include these activities in their daily schedules. Across 21 simulated days, each agent successfully generates all mandatory travel-related activities without omission. This result indicates that the agents reliably interpret their assigned profiles and consistently generate the corresponding required activities.

\begin{figure}[!h]
    \centering
    \subfloat[][Grocery shopping]{\includegraphics[height=8cm]{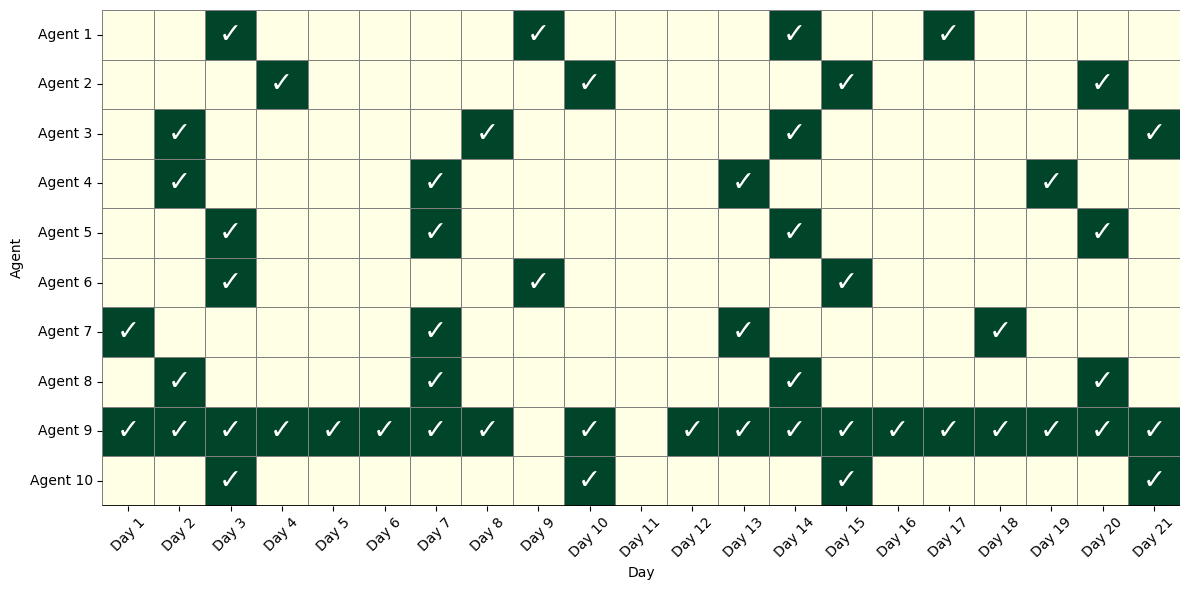}}\\
    \subfloat[][Recreational outings]{\includegraphics[height=8cm]{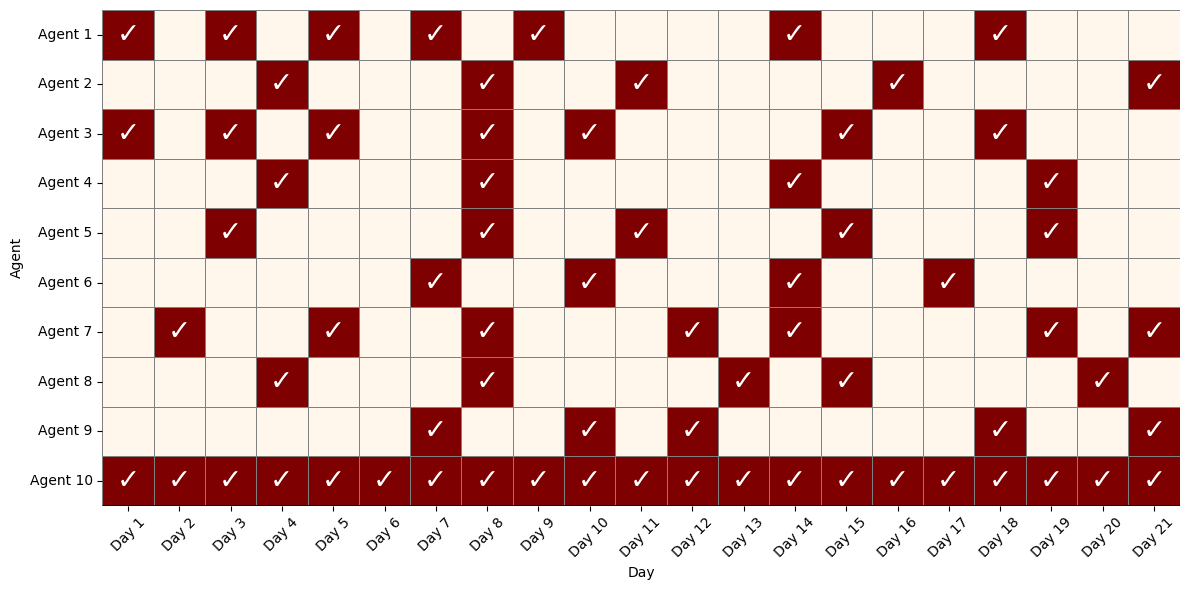}}
    \caption{Agents' daily generated activities over 21 days of simulation. Boxes with checkmarks indicate that the activities noted in the figure are scheduled by the corresponding agent on that day.}
    \label{fig:agent_activity_generation}
\end{figure}

For non-mandatory activities, we qualitatively evaluate the agents’ ability to generate maintenance (exemplified by grocery shopping) and discretionary activities (exemplified by recreational outings) based on household needs. Real-world households periodically perform maintenance activities to fulfill operational requirements and discretionary activities to support household well-being. An LLM agent should recognize such needs by referencing recent activity patterns and generate relevant activities accordingly. The occurrence of grocery shopping and recreational activities generated by each agent over the 21-day simulation is shown in \Cref{fig:agent_activity_generation}.

As shown in \Cref{fig:agent_activity_generation}, the LLM agents are capable of generating both types of non-mandatory activities in reasonable and context-sensitive patterns. For grocery shopping (shown in \Cref{fig:agent_activity_generation}a), most agents perform the activity periodically, averaging 1.3 times per week. The shortest interval between grocery runs is 3 days, and the longest is 7 days, both plausible for a typical suburban household. Furthermore, agent 9, which is configured to value fresh groceries more highly, conducts grocery shopping 19 times over 21 days. This outcome suggests that the agent successfully interprets profile-level preferences and adjusts its behavior accordingly. Similarly, \Cref{fig:agent_activity_generation}b demonstrates that LLM agents can also generate recreational outings with realistic frequency and variability. For agents without elevated recreational preferences, the average number of recreational activities generated is 1.8 times per week, with intervals between events ranging from 2 to 6 days. These numbers are plausible for a typical household. Additionally, agent 10, whose profile emphasizes a strong preference for outdoor exercise, schedules recreational activities every day, demonstrating its ability to incorporate profile-based preferences into activity generation. Overall, these results show that LLM agents can recognize household needs, infer appropriate timing, and generate activities that align with the agent profiles.

\subsubsection{Formatting travel plans}

\begin{figure}[h!]
  \centering
  \includegraphics[width=0.9\textwidth]{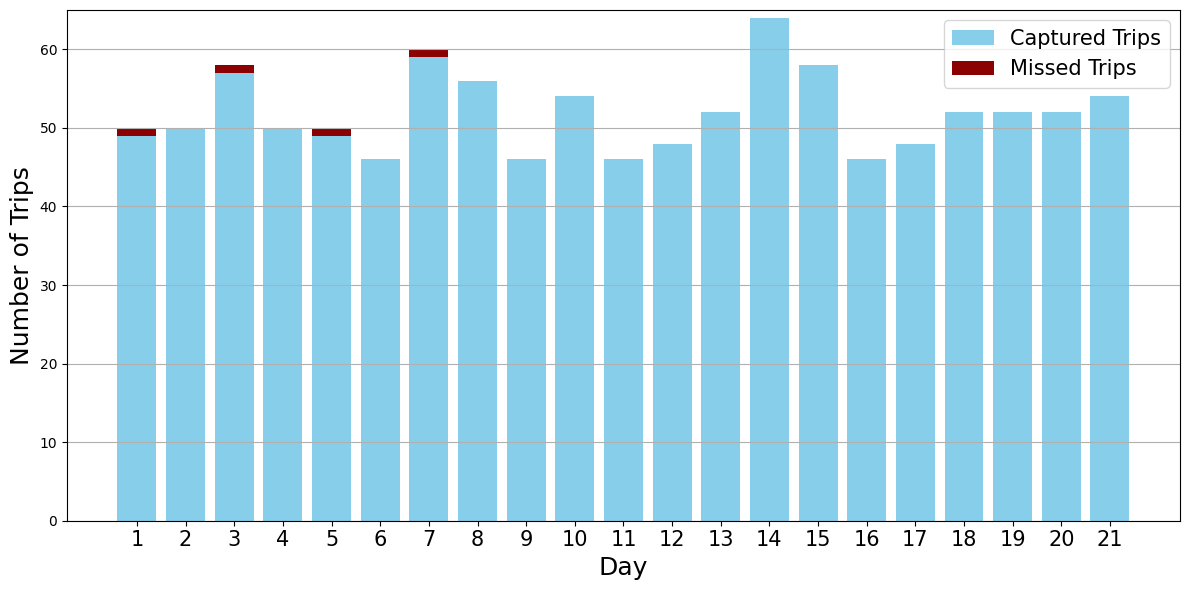}
  \caption{LLM agents' performance in formatting travel plans over 21 days of simulation}\label{fig:example_miss_trips}
\end{figure}

To bridge the gap between the LLM agents' natural language outputs and the structured input requirements of DTALite, each agent is instructed to convert its travel plans into the required structured format. This conversion process introduces potential sources of error, particularly if any trips are omitted or misrepresented during the transition from natural language to structured data. To evaluate the reliability of this process, we compare the number of trips described in the agents' natural language plans with those successfully captured in the structured outputs. The comparison is summarized in \Cref{fig:example_miss_trips}.

As shown in \Cref{fig:example_miss_trips}, the LLM agents demonstrate strong and consistent performance in converting their travel plans into the required structured format. Out of a total of 1,092 trips generated over 21 days, only 4 trips (0.4\%) are omitted during formatting. This error rate is minimal and suggests that the agents are able to understand formatting instructions and correctly represent their intended trips in structured form. Notably, most of the missed trips occurred during the first week of the simulation, while no formatting errors are observed in the last two weeks. This pattern suggests that the reliability of the process may improve over time, possibly as the agents accumulate more experience with plan generation and structure formatting.

\subsubsection{Learning and adjustment}

\begin{figure}[!h]
    \centering
    \subfloat[][Dad's morning commute arrival times]{\includegraphics[height=8.25cm]{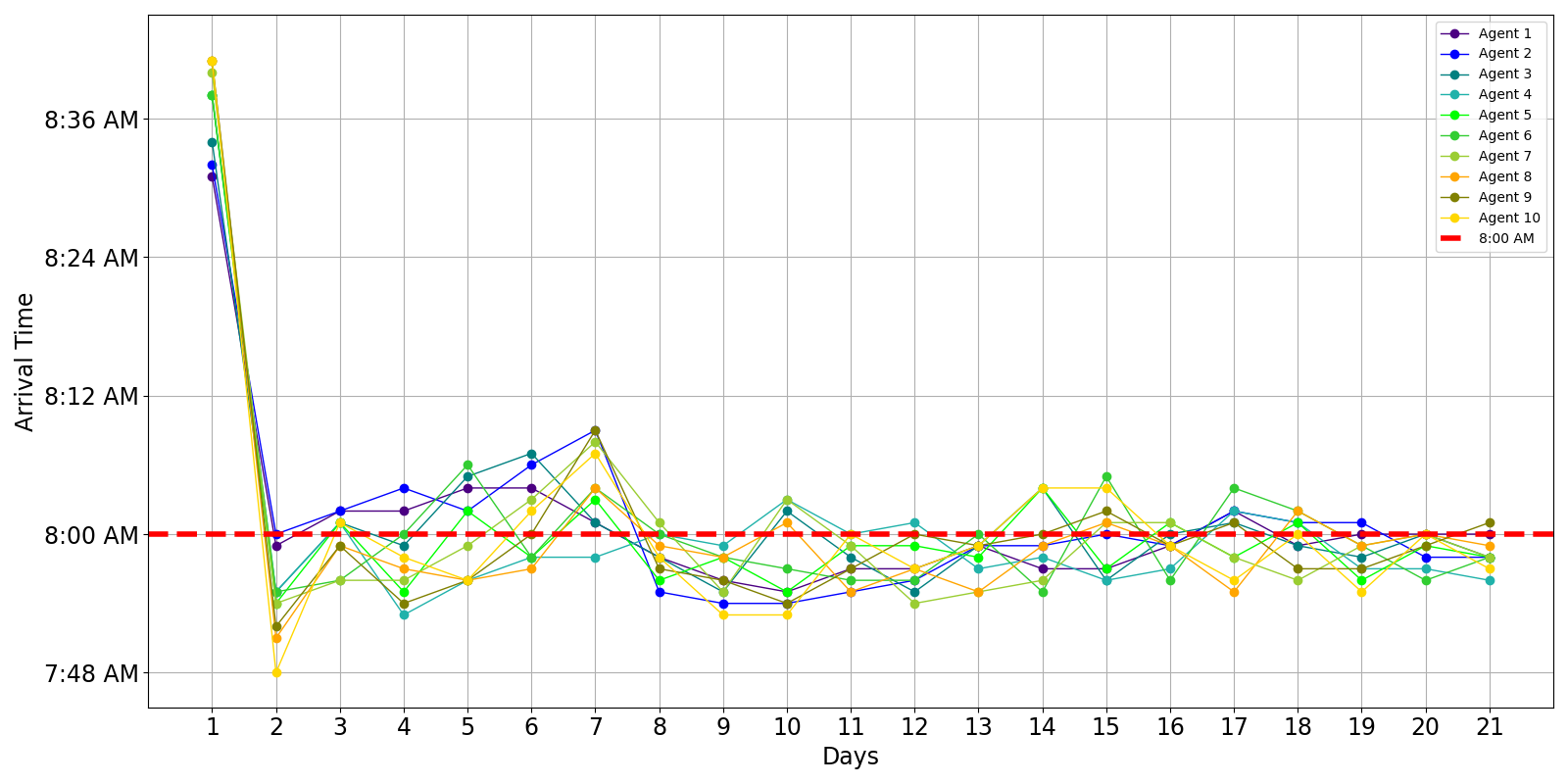}}\\
    \subfloat[][Kid's morning commute arrival times]{\includegraphics[height=8.25cm]{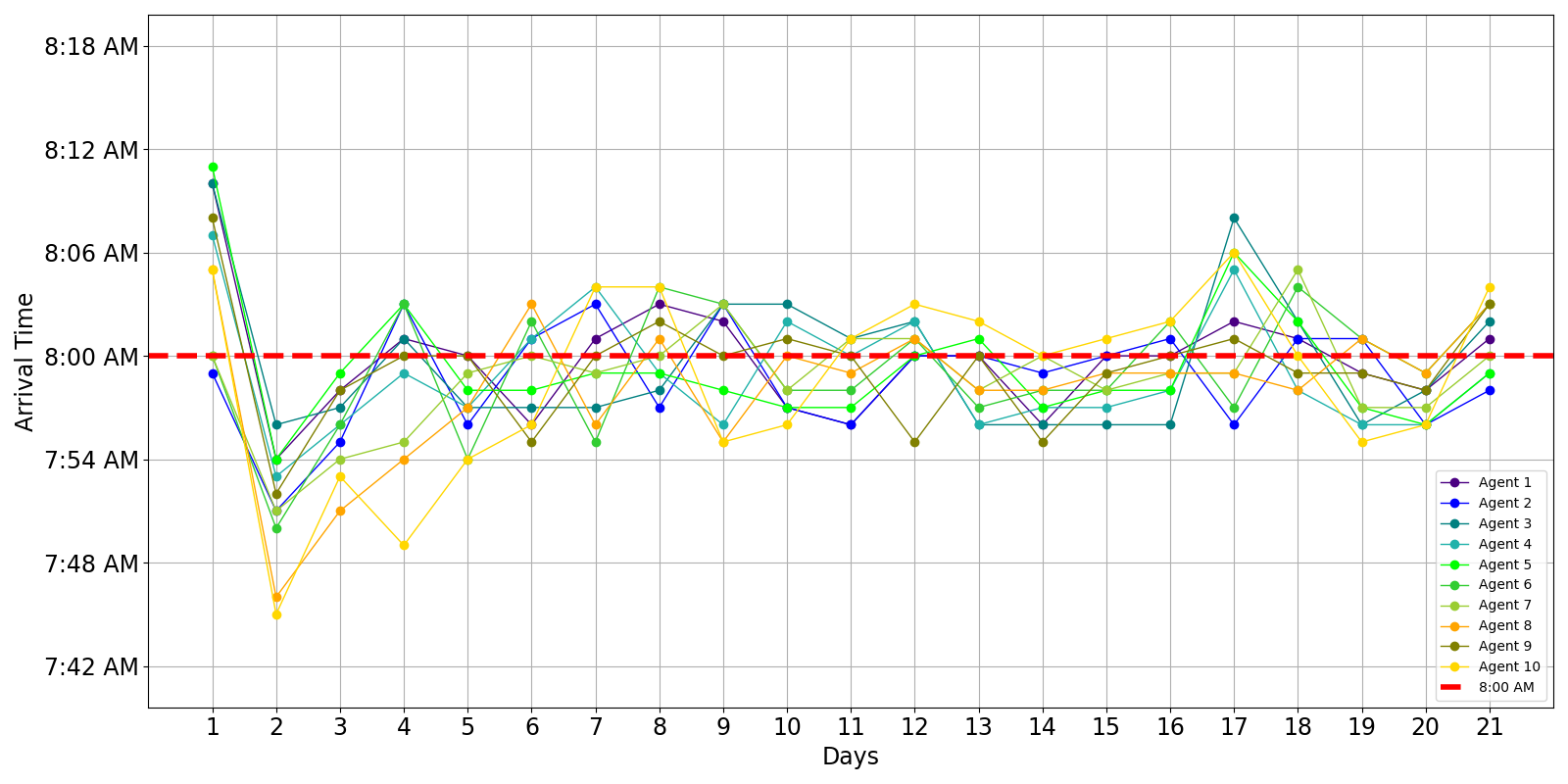}}
    \caption{Morning commute arrival times for the agents}
    \label{fig:agent_arrival_times}
\end{figure}

To assess the agents’ ability to learn from experience and adapt their travel plans through interactions with the transportation infrastructure, we examine the patterns and evolution of their morning commutes. Since both the dad and kid in each household must commute to work or school with fixed arrival time constraints, we can observe how agents adapt their strategies as the simulation progresses. The morning commute arrival times for dads and kids of every household are shown in \Cref{fig:agent_arrival_times}.

As illustrated in \Cref{fig:agent_arrival_times}, the agents are able to learn from previous travel outcomes and adjust their departure times to better meet the arrival time requirements while avoiding excessively early or late arrivals. This is evident in both panels: dads' commute times in \Cref{fig:agent_arrival_times}(a) and kids' commute times in \Cref{fig:agent_arrival_times}(b). For dads' commutes, agents initially arrive significantly late on Day 1. However, they quickly adapt to the situation. Most agents adjust their departure times by day 2 to ensure arrival before the 8:00 AM work start time. Throughout the simulation, agents continue to make incremental adjustments in response to feedback, gradually improving the timing of their arrivals. By day 21, most agents arrive right around 8:00 AM, demonstrating a marked improvement from their initial strategies. A similar pattern is observed for the kids’ morning commutes, although the magnitude of adjustment is smaller. Agents initially arrive late to school, then compensate by arriving early on subsequent days. Over time, they refine their behavior and settle into an arrival time pattern that is close to the 8:00 AM school start time. These patterns collectively indicate that the agents can evaluate outcomes, adjust strategies, and progressively optimize their travel behavior based on experience.

\section{Challenges and Near-term Strategy} \label{sec:challenges}
Our previous discussion and demonstration illustrated the potential, advantages, and feasibility of the LLM-agent-based modeling framework. However, there are still key challenges in the development and implementation of this framework. In this section, we outline some of these challenges and future research directions, as well as discuss a hybrid modeling framework that can serve as a near-term development strategy.

\subsection{Challenges and research directions} \label{subsec:future_challenges_directions}
\subsubsection{Enhancing behavioral alignment}
While existing research on LLMs and LLM agents has demonstrated their strong potential in advancing ABMs of transportation systems, several methodological challenges remain. Addressing these challenges is essential for improving the behavioral fidelity and applicability of LLM agents in transportation system ABMs. Below, we outline some key areas requiring further research:

\begin{itemize}
    \item \textbf{Modeling the randomness in human behavior}: Human travel behavior exhibits inherent randomness, as individuals may make different choices even when faced with the same context. Traditional modeling frameworks address this variability through stochastic models grounded in random utility theory, which help capture the stochasticity of humans in decision-making. However, LLMs may struggle to replicate this randomness, as they are designed to predict the most probable next word or sequence based on training data, often resulting in uniform and repetitive responses. This limitation has been observed in social science \citep{dominguez2024questioning,park2024diminished} and economics research \citep{chen2023emergence,korinek2023language}. While LLM hyperparameters, such as the ``temperature" setting, can modulate randomness in text generation, studies have shown that temperature adjustments alone are insufficient to achieve the level of behavioral variability observed in human travelers \citep{wang2024large,liu2024can}. Further research is needed to address this misalignment and develop strategies to better capture randomness in LLM-agent-based travel behavior simulation.

    \item \textbf{Understanding and reducing inherent bias in LLMs}: Despite being trained on extensive and diverse datasets, LLMs are not immune to biases embedded in their training data. These biases may lead to systematic distortions in the representation of human travel behavior, particularly in the misrepresentation of certain population groups. This issue has been documented in existing research as well \citep{wang2025large}, highlighting the need for a comprehensive evaluation of biases in LLM-agent-generated travel behavior. Future research need to work on identifying and quantifying biases in LLM-based travel behavior simulations and developing robust mitigation strategies to ensure fair and accurate representation across different travelers.

    \item \textbf{Incorporating attitude variables}: Existing research on LLM agents has primarily focused on profiling agents through socio-demographic attributes and memory systems. However, travel behavior studies suggest that incorporating attitude variables can improve the accuracy of behavioral predictions \citep{mokhtarian2024pursuing}. Similarly, recent work in LLM-based behavioral modeling shows that attitudinal inputs can enhance the realism and depth of agent representations \citep{chuang2024beyond}. Future studies should explore how to effectively integrate attitudinal variables into LLM-based agent profiles, including identifying data sources and building methods and procedures for this integration.

    \item \textbf{Realigning values for travel behavior representation}: The training process of LLMs often involves value alignment to ensure that models adhere to socially desirable norms, thereby preventing the generation of harmful, offensive, or inappropriate content. However, this alignment process may inadvertently introduce biases that overemphasize certain value-driven behaviors in ways that do not accurately reflect real-world human behavior. Empirical studies have identified similar distortions in value-laden decision-making and interactions of LLMs \citep{strachan2024testing,liu2024can}. Future research could investigate emerging techniques such as LLM unlearning \citep{yao2024large,liu2025rethinking} to realign travel behavior representations while maintaining ethical and responsible AI deployment in transportation modeling contexts.
    
\end{itemize}

\subsubsection{Improve simulation scalability}
Another major challenge in LLM-agent-based modeling of transportation systems is scalability. Transportation systems inherently consist of a large number of interacting agents, each responsible for making various travel decisions. Scaling an LLM-agent-based framework to accommodate these interactions presents significant challenges, primarily due to computational resource constraints and processing latency. The first major obstacle is computational power. LLMs, with billions of parameters, require substantial computing resources ranging from GPUs to TPUs to operate efficiently. As the number of agents increases, memory consumption and computational demands scale proportionally, often exceeding the capacity of standard computational devices. The second obstacle is processing latency. Unlike traditional ABMs, which employ simplified decision rules or lightweight machine learning models, LLMs perform computationally intensive inference. Each agent decision involves token-based computations and probabilistic reasoning, resulting in significant delays per simulation step. When simulations involve a large number of agents across multiple iterations, this latency accumulates, posing a substantial challenge to large-scale applications.

Advancements in LLM technology have introduced methods to mitigate scalability constraints in LLM-agent-based modeling and simulation systems. At the LLM core level, techniques such as quantization \citep{liu2023llm}, PagedAttention \citep{kwon2023efficient}, batching, and caching can optimize computational efficiency by approximating and streamlining the inference process. At the simulation pipeline level, approaches such as representative agents \citep{chopra2024limits} and meta-prompting \citep{yan2024opencity} can reduce the number of agents required and simplify prompt structures, thereby minimizing computational overhead and latency. However, transportation system simulations require not only large populations but also multiple iterations to achieve stable and reliable behavioral outcomes. Ensuring behavioral realism remains critical, necessitating a balance between computational efficiency and the accuracy of agent behavior representation. Future research should explore innovative methodologies to optimize the trade-off between scalability and fidelity, enabling efficient yet behaviorally accurate LLM-agent-based ABMs for large-scale transportation systems.

\subsubsection{Advancing model validation}

The final main challenge for the LLM-agent-based modeling framework is validation. Although initial efforts suggest that LLMs can replicate human travel behavior effectively in certain scenarios and generate plausible learning and adjustments, extensive validation is essential to assess their behavioral accuracy across a broader range of tasks and contexts. For instance, it remains unclear how precisely LLM agents emulate human cognitive processes related to travel behavior, particularly regarding how individuals process information, perceive travel conditions, and adaptively respond to changing environments in real-time. To address these uncertainties, detailed experimental validation is necessary to benchmark LLM-generated decisions against empirical observations of real-world traveler behaviors. Moreover, achieving behavioral fidelity at the individual decision-making level does not inherently guarantee that system-wide simulation outcomes will correspond closely to real-world patterns, trends, or aggregate system behaviors. It is possible for inaccuracies at the individual level to propagate, potentially magnifying errors at the system level. Consequently, an extensive and systematic validation process must incorporate validation at multiple scales---both at the micro-level of individual traveler decisions and the macro-level of system performance. This multi-level validation process would involve quantitative assessments comparing simulation outputs with empirical data, including observed traffic flows, modal splits, and congestion patterns. Addressing these validation challenges will require both methodological advances and the development of comprehensive, representative datasets and benchmarks, which should be a direction of future research.

\subsection{Hybrid modeling: a near-term strategy}

\begin{figure}[h!]
  \centering
  \includegraphics[width=0.9\textwidth]{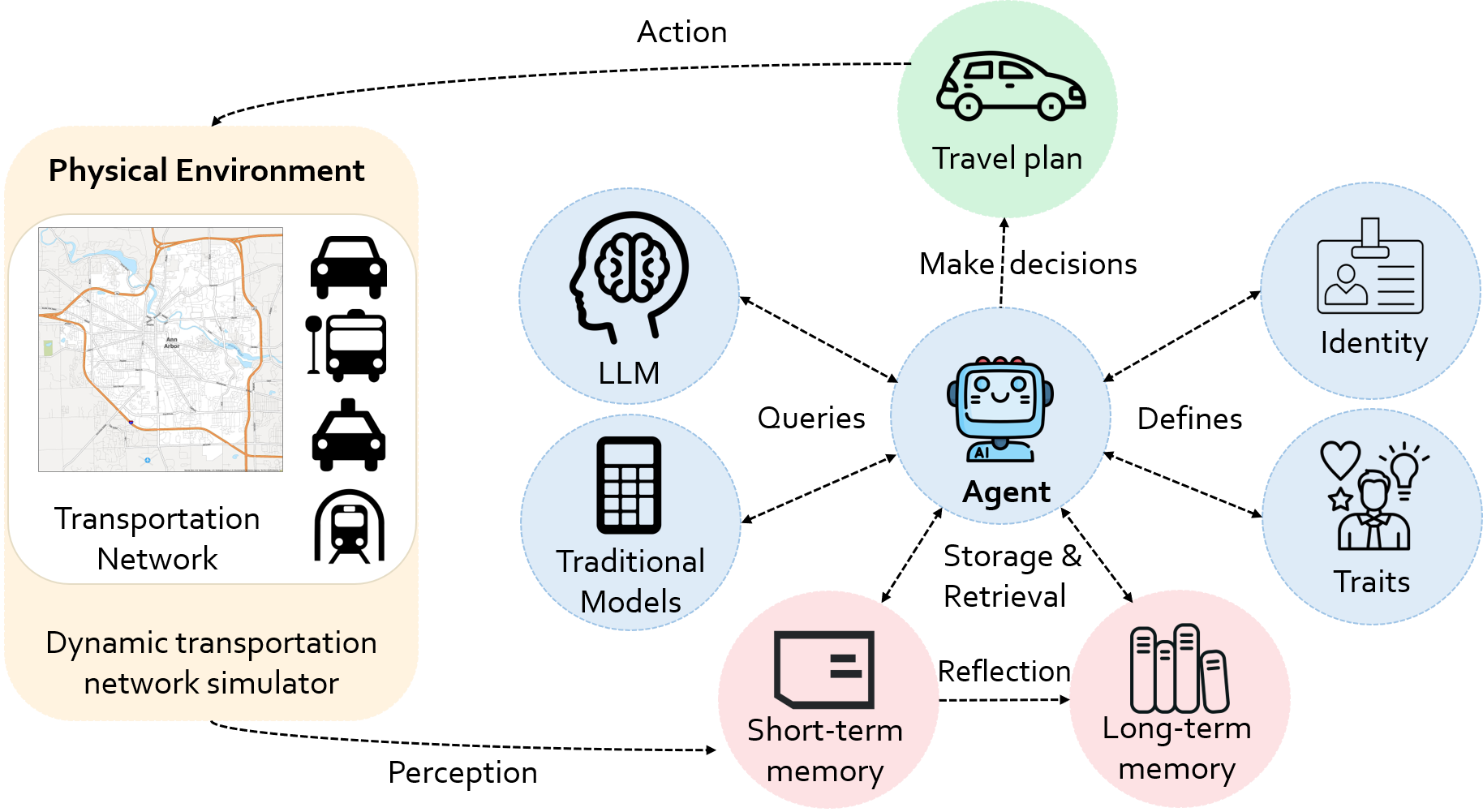}
  \caption{Overview of the hybrid modeling framework}\label{fig:hy_system_workflow}
\end{figure}

As discussed in \Cref{subsec:future_challenges_directions}, the LLM-agent-based modeling framework for transportation systems continues to face challenges related to behavioral alignment and scalability. Addressing both challenges will likely require long-term research and technological innovations. As a near-term solution to facilitate the integration of LLM agents into transportation system modeling and simulation, we propose a hybrid modeling approach (illustrated in \Cref{fig:hy_system_workflow}) that combines LLM-based agents with existing modeling techniques. This approach seeks to leverage the strengths of LLMs---including their natural language reasoning, contextual adaptation, and generalization capabilities---while mitigating their current limitations and uncertainties through established methodologies in agent-based modeling and behavior science.

In the integrated modeling framework, LLM agents and traditional modeling methods each occupy distinct, complementary roles within the simulation pipeline, capitalizing on their respective strengths:

\begin{itemize}
    \item \textbf{LLM agents} are well-suited for tasks that are inherently unstructured and difficult to formalize mathematically. Such tasks often lack standardized quantitative formulations that generalize well across varied contexts. They also pose calibration challenges, as the underlying behavioral parameters are typically implicit, high-dimensional, and difficult to specify using traditional modeling techniques. LLM agents help mitigate these limitations by leveraging flexible reasoning and language-based representations, and they can effectively handle tasks such as constructing detailed daily schedules and tours under restrictions, resolving conflicts among overlapping activity priorities, and dynamically adjusting plans based on accumulated experiences.

    \item \textbf{Traditional models}, by contrast, excel at tasks that are structured, quantifiable, and well-supported by established discrete choice theory, such as mode choice and route assignment. These tasks involve clearly defined behavioral parameters, benefit from mature mathematical formulations, and are supported by standardized calibration pipelines. Their ability to scale efficiently across large populations makes them a valuable tool in the hybrid framework, and they can complement the capabilities of LLM agents by reinforcing accuracy and consistency in the structured components of the simulation pipeline.
\end{itemize}

The hybrid modeling approach aims to gradually integrate LLMs into existing transportation modeling pipelines. By strategically combining these methodologies, this approach offers an adaptive near-term solution that bridges LLM-driven innovation with the established foundations of agent-based models. Beyond the integration process of LLMs, the hybrid approach can also be leveraged in the long term for resource-constrained modelers and practitioners.

\section{Conclusion} \label{sec:conclusion}
In this paper, we introduce an LLM-agent-based modeling framework for transportation systems, leveraging the natural language understanding, reasoning, and autonomous decision-making capabilities of LLM agents to serve as adaptive traveler agents within an ABM framework. We argue that LLM agents have the potential to address key challenges in transportation ABMs, particularly in relaxing behavioral assumptions, improving data efficiency, enhancing model flexibility, and facilitating better learning and adjustment behavior of agents. Our proposed framework integrates LLM agents with dynamic transportation network simulators, using LLM agents as proxies for human travelers. The framework is designed to align closely with human behavioral processes, incorporating a memory system that mimics human recall, an information retrieval mechanism, and a decision-making process that enables agents to learn from prior experiences and adjust their travel behavior dynamically. The agent's decision-making flow is aligned closely with the activity-based travel demand models to ensure behavioral realism. Furthermore, we present evidence from established literature demonstrating that LLM agents can exhibit human-like travel behavior in activity generation and scheduling; travel choice; and adaptive learning. We then present a proof-of-concept simulation of our proposed LLM-agent-based modeling framework. The simulation results shows that the proposed LLM-agent-based framework demonstrates promising capabilities in generating context-aware travel behavior. These results provide initial evidence that LLM agents can serve as behaviorally rich traveler proxies, supporting the feasibility and practicality of integrating LLMs into transportation ABMs.

Despite its promise, several challenges remain. Ensuring behavioral alignment between LLM agents and human travelers, particularly in capturing variability and randomness in decision-making and mitigating underlying biases in LLMs, remains an open research question. Additionally, scalability constraints present limitations that must be addressed to enable large-scale simulations. Lastly, extensive validation of both individual agent behavior and system-wide outcomes will be essential for broader adoption. As a near-term solution, we propose a hybrid modeling approach that integrates LLM agents with traditional modeling techniques. Future research should focus on enhancing behavioral fidelity, improving computational efficiency, refining hybrid modeling techniques, and establishing robust evaluation methods to further advance the LLM-agent-based modeling paradigm for transportation systems.

\section*{Acknowledgement}
The work described in this paper was partly supported by research grants from the National Science Foundation (CMMI-2233057, CMMI-2240981). The authors would like to thank Manzi Li, Minghui Wu, Xi Lin, Yingnan Yan, and Zhichen Liu for their contribution during the early stages of this work.

\section*{Declaration of generative AI and AI-assisted technologies in the writing process}
During the preparation of this work, the authors used ChatGPT to improve language
clarity and readability. After using this tool/service, the authors reviewed and edited the content as needed and take full responsibility for the content of the published article.

\newpage
\bibliographystyle{apalike}

\end{document}